\documentclass{article}

\usepackage[english]{babel}
\usepackage[utf8x]{inputenc}
\usepackage[T1]{fontenc}

\usepackage{blindtext}

\usepackage{amsfonts, amsmath, amssymb, amsthm}
\usepackage{yfonts} 
\usepackage{gensymb} 

\usepackage{physics}    

\usepackage{xpatch} 

\usepackage{mathtools} 

\usepackage{mathalpha}
\usepackage{bm}
\usepackage{verbatim}
\usepackage{euscript}
\usepackage{graphicx} 

\usepackage{paralist}
\usepackage{cancel} 

\usepackage{bbm} 

\usepackage{nicefrac}

\usepackage{enumitem}
\usepackage{graphicx}
\usepackage{soul}
\usepackage{float}

\usepackage{xspace}
\usepackage{wrapfig}

\usepackage{subcaption}
\usepackage{caption} 

\usepackage[usenames,dvipsnames]{color}
\usepackage[dvipsnames]{xcolor}

\usepackage{csvsimple}
\usepackage{etoolbox}  
\usepackage{pgf}

\usepackage[colorinlistoftodos]{todonotes}
\usepackage[colorlinks=true, allcolors=black]{hyperref} 
\hypersetup{
  colorlinks   = true, 
  urlcolor     = blue, 
  linkcolor    = blue, 
  citecolor   = blue 
}
\usepackage[noabbrev,capitalize,nameinlink]{cleveref}

\usepackage{natbib}

\usepackage{booktabs} 
\usepackage{tabls} 
\usepackage{array}
\usepackage{longtable}

\usepackage[normalem]{ulem}
\usepackage{soul}  

\usepackage{pifont}

\usepackage{array} 

\DeclareMathOperator*{\argmax}{arg\,max}

\DeclareMathSymbol{\shortminus}{\mathbin}{AMSa}{"39}
\makeatletter
\newcommand{\shorteq}{%
  \settowidth{\@tempdima}{--}
  \resizebox{\@tempdima}{\height}{=}%
}
\makeatother

\newcommand{\cardinality}[1]{ \left\lvert #1 \right\rvert }

\newcommand{\prob}[1]{\mathbb{P} \left\{ #1 \right\}}

\newcommand{\probunder}[2]{\mathbb{P}_{#1} \left\{ #2 \right\}}

\newcommand{\expectation}[1]{\mathbb{E} \left[ #1 \right]}

\newcommand{\expund}[2]{\mathbb{E}_{#1} \left[ #2 \right]}

\newcommand{\gaussian}[2]{\mathcal{N} \left(#1, #2\right)}

\newcommand{\nonnegint}{\mathbb{Z}_{ \geq 0}}

\newcommand{\appxbigO}[1]{\widetilde{O} \left( #1 \right)}

\newcommand{\numberthis}{\addtocounter{equation}{1}\tag{\theequation}}

\newtheorem{lemma}{Lemma}
\newtheorem{claim}[lemma]{Claim}
\crefname{claim}{Claim}{Claims}

\crefname{axiom}{Axiom}{Axioms}

\theoremstyle{definition}

\newtheorem{definition}{Definition}
\newtheorem{property}[definition]{Property}
\crefname{property}{Property}{Properties}

\crefname{philosophy}{Philosophy}{Philosophies}

\crefname{assumption}{Assumption}{Assumptions}

\usepackage{thmtools}
\usepackage{thm-restate} 

\makeatletter
\newcommand{\saveequation}[2]{
  #2 \label{#1}
  \protected@write\@mainaux{}{\string\SAVEEQUATION{#1}{\unexpanded{\unexpanded{#2}}}}%
}
\newcommand{\SAVEEQUATION}[2]{%
  \global\@namedef{SAVEDEQUATION@#1}{#2}%
}
\newcommand{\repeatequation}[1]{%
  \ifcsname SAVEDEQUATION@#1\endcsname
    \@nameuse{SAVEDEQUATION@#1}\tag{\ref{#1}}%
  \else
    ?? \notag
  \fi
}
\makeatother

\newenvironment{remark-nolabel}{\noindent{\bf Remark}\hspace*{1em}}{\bigskip}

\newcolumntype{L}[1]{>{\raggedright\let\newline\\\arraybackslash\hspace{0pt}}m{#1}}
\newcolumntype{C}[1]{>{\centering\let\newline\\\arraybackslash\hspace{0pt}}m{#1}}
\newcolumntype{R}[1]{>{\raggedleft\let\newline\\\arraybackslash\hspace{0pt}}m{#1}}

\definecolor{rowgray}{RGB}{245,245,245} 
\definecolor{rowwhite}{RGB}{255,255,255}

\newcommand{\Real}{\mathbb{R}}

\newcommand{\muhat}{\widehat{\mu}}

\long\def\comment#1{}

\newcommand{\rama}[1]{}

\newcommand{\bu}[1]{\underline{\textbf{#1}}}

\newcounter{relctr} 
\counterwithin*{relctr}{section} 

\crefname{relctr}{step}{steps}
\Crefname{relctr}{Step}{Steps}
\creflabelformat{relctr}{#2(#1)#3}

\newcommand\labelrel[2]{%
  \begingroup
    \refstepcounter{relctr}%
    \stackrel{\textnormal{(\alph{relctr})}}{\mathstrut{#1}}%
    \originallabel{#2}%
  \endgroup
}
\AtBeginDocument{\let\originallabel\label} 

\newcommand{\handout}[6]{
   \renewcommand{\thepage}{\arabic{page}} 
   \noindent
   \begin{center}
   \framebox{
      \vbox{
    \hbox to 5.78in { {\bf #2}
     	 \hfill #3 }
       \vspace{4mm}
       \hbox to 5.78in { {\Large \hfill #6  \hfill} }
       \vspace{2mm}
       \hbox to 5.78in { {\it #4 \hfill #5} }
      }
   }
   \end{center}
   \vspace*{4mm}
}

\makeatletter
\let\iftwocolumn\if@twocolumn
\makeatother
\crefname{property}{Property}{Properties}
\Crefname{property}{Property}{Properties}

\usepackage{algorithmicx, algpseudocode}
\usepackage[ruled]{algorithm}

\usepackage{algorithm}
\usepackage{algpseudocode}

\usepackage{etoolbox}
\usepackage{lipsum}

\makeatletter
\newcommand*{\algrule}[1][\algorithmicindent]{%
  \hspace*{.2em}
  \vrule 
  \hspace*{\dimexpr#1-.2em-.4pt}%
}

\newcommand{\StatePar}[1]{%
  \State\parbox[t]{\dimexpr\linewidth-\ALG@thistlm}{\strut #1\strut}%
}
\renewcommand{\ALG@beginalgorithmic}{\offinterlineskip}

\newcount\ALG@printindent@tempcnta
\def\ALG@printindent{%
  \ifnum \theALG@nested > 0
    \ifx\ALG@text\ALG@x@notext
    \else
      \unskip
      \ALG@printindent@tempcnta=1
      \loop
        \algrule[\csname ALG@ind@\the\ALG@printindent@tempcnta\endcsname]%
        \advance \ALG@printindent@tempcnta 1
        \ifnum \ALG@printindent@tempcnta<\numexpr\theALG@nested+1\relax
      \repeat
        \fi
    \fi
}
\patchcmd{\ALG@doentity}{\noindent\hskip\ALG@tlm}{\ALG@printindent}{}{\errmessage{failed to patch}}
\makeatother

\algrenewcommand\algorithmicend{\strut\textbf{end}}
\algrenewcommand\algorithmicdo{\strut\textbf{do}}
\algrenewcommand\algorithmicwhile{\strut\textbf{while}}
\algrenewcommand\algorithmicfor{\strut\textbf{for}}
\algrenewcommand\algorithmicforall{\strut\textbf{for all}}
\algrenewcommand\algorithmicloop{\strut\textbf{loop}}
\algrenewcommand\algorithmicrepeat{\strut\textbf{repeat}}
\algrenewcommand\algorithmicuntil{\strut\textbf{until}}
\algrenewcommand\algorithmicprocedure{\strut\textbf{procedure}}
\algrenewcommand\algorithmicfunction{\strut\textbf{function}}
\algrenewcommand\algorithmicif{\strut\textbf{if}}
\algrenewcommand\algorithmicthen{\strut\textbf{then}}
\algrenewcommand\algorithmicelse{\strut\textbf{else}}

\algrenewcommand\algorithmicrequire{\strut\textbf{Input:}}
\algrenewcommand\algorithmicensure{\strut\textbf{Output:}}

\let\oldState\State
\renewcommand{\State}{\oldState\strut}

\algnewcommand{\LineComment}[1]{\State \(\triangleright\) #1}

\usepackage[main,final]{neurips_2026}

\usepackage[T1]{fontenc}    
\usepackage{hyperref}       
\usepackage{url}            
\usepackage{booktabs}       
\usepackage{amsfonts}       
\usepackage{nicefrac}       
\usepackage{microtype}      
\usepackage{xcolor}         

\title{On the Impact of Extra Information in Bandit Learning: More is Not Always Merrier}

\author{%
  Ramakrishnan Krishnamurthy \\
  Courant Institute of Mathematical Sciences, New York University
}

\begin{document}

\maketitle

\newcommand{\alg}{\textsc{Alg}}
\newcommand{\ucbalg}{\textsc{Ucb1}}
\newcommand{\nondegalg}{\textsc{Meta}}
\newcommand{\ucb}{\textsc{Ucb}}
\newcommand{\ts}{\textsc{Ts}}
\renewcommand{\etc}{\textsc{Etc}}

\newcommand{\boldzero}{\mathbf{0}}
\renewcommand{\dd}{\textsc{D}}
\newcommand{\ee}{\textsc{E}}
\newcommand{\ff}{\textsc{F}}

\newcommand{\side}{\mathcal{D}}
\newcommand{\actions}{\mathcal{A}}
\newcommand{\inst}{\mathcal{I}}
\newcommand{\instfix}{\inst^{\text{fixed}}}
\newcommand{\hist}{\mathcal{H}}

\newcommand{\mub}{\boldsymbol{\mu}}

\newcommand{\istar}{\widehat{i_\star}}
\newcommand{\pd}[1]{\mathbb{P}_{\dd} \left\{ #1 \right\}}
\newcommand{\pe}[1]{\mathbb{P}_{\ee} \left\{ #1 \right\}}

\newcommand{\exd}[1]{\expund{\dd}{ #1}}
\newcommand{\exe}[1]{\expund{\ee}{ #1}}

\newcommand{\vd}[1]{\mathbb{V}_{\dd} \left[ #1 \right]}
\newcommand{\ve}[1]{\mathbb{V}_{\ee} \left[ #1 \right]}

\begin{abstract}
More data is commonly assumed to improve learning performance, however, 
we show that this intuition can surprisngly fail in sequential decision-making. 
Considering a stochastic bandits setup with exogenous extra information, we ask if the free and unbiased extra reward samples are guaranteed to reduce the regret of the learner.
We introduce a formal notion of `non-degradation under extra information' and demonstrate that the widely used Upper Confidence Bound algorithm ($\ucb$) violates this property, i.e.,
there exist benign extra-information under which $\ucb$ incurs strictly higher regret than when run without extra information.
We contrast this behaviour with Thompson Sampling and Explore-Then-Commit, which appear to benefit from the same extra information.
Constructively, we propose a meta-algorithm that provably enforces non-degradation for a broad class of bandit algorithms.
We also run empirical simulations to study our meta-algorithm.
\end{abstract}
\section{Introduction}

In classical supervised learning with i.i.d.\ data, more data almost always helps performance \citep{vapnik1999overview,shalev2014understanding}.
Learning curves formalize this principle by quantifying how risk, accuracy, or error rates improve as the number of samples increases.
Under standard assumptions, additional unbiased data monotonically improves estimation and generalization \citep{hastie2009elements}.

Sequential decision-making problems such as reinforcement learning and multi-armed bandits depart fundamentally from this paradigm.
The learner does not passively observe data, but must \emph{act} to generate observations \citep{sutton1998reinforcement}.
Actions that are suboptimal incur immediate regret, creating an exploration-vs-exploitation trade-off.
As a result, algorithmic performance is measured not only by the quality of the final model, but by cumulative reward obtained throughout the learning process \citep{lai1985asymptotically}.

This distinction raises a basic and surprisingly underexplored question:
\begin{quote}
\emph{Does extra data always improve performance in
sequential decision-making?}
\end{quote}

In this paper, we show that the answer is no.
Even when extra data consists of unbiased, non-adversarial samples from
the environment, and is free (incurrs no regret), providing it naively can \emph{increase} overall regret.
Unlike in supervised learning, where more data uniformly sharpens estimates,
in online learning, additional data can distort exploration incentives
and degrade performance.

\begin{figure*}[ht]
    \centering
    \begin{subfigure}{.30\linewidth}
        \centering
        \includegraphics[width=\linewidth]{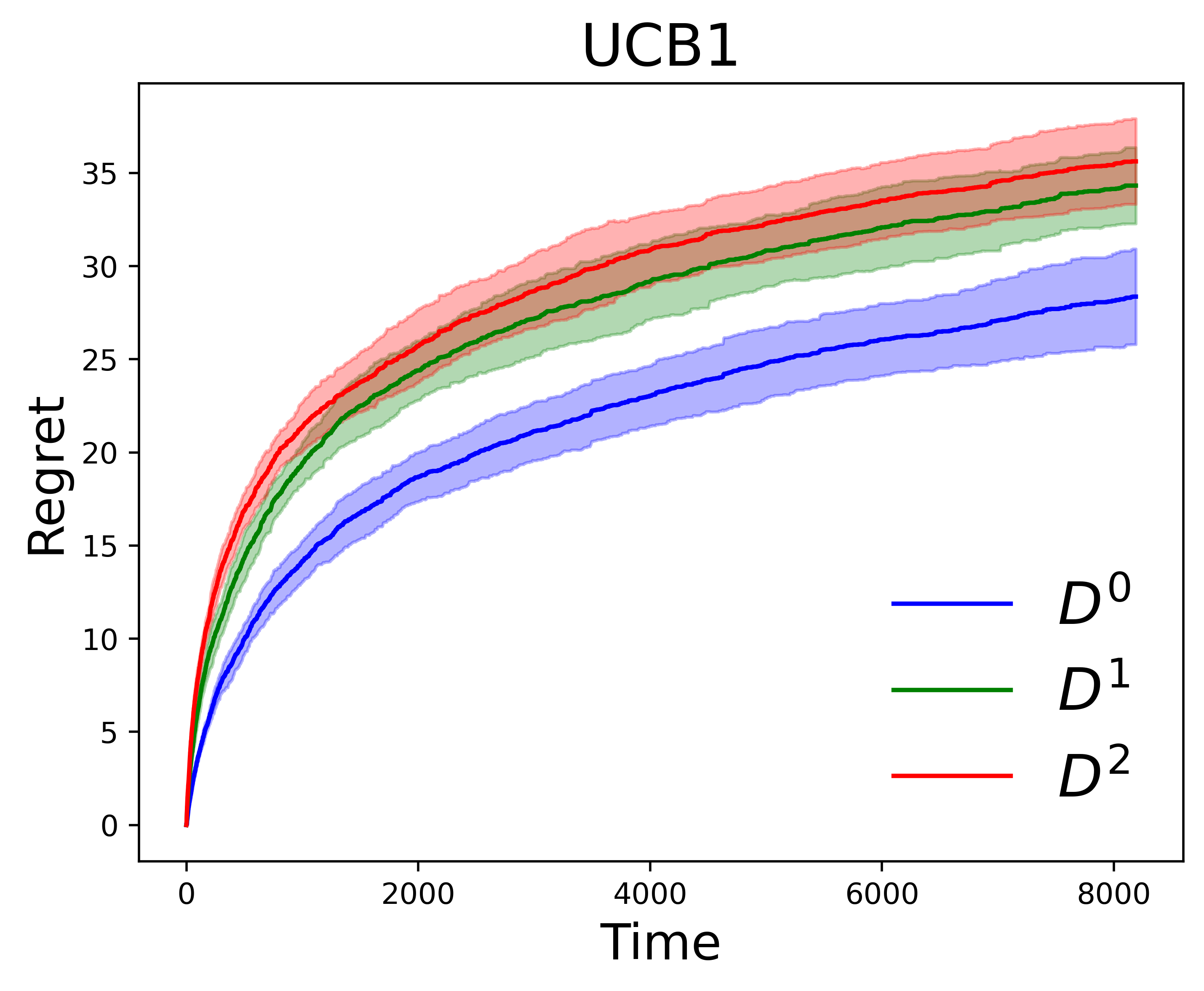}
        \caption{$\ucb$ with extra info.}
        \label{fig:ucb-side-info}
    \end{subfigure}%
    \begin{subfigure}{.30\linewidth}
        \centering
        \includegraphics[width=\linewidth]{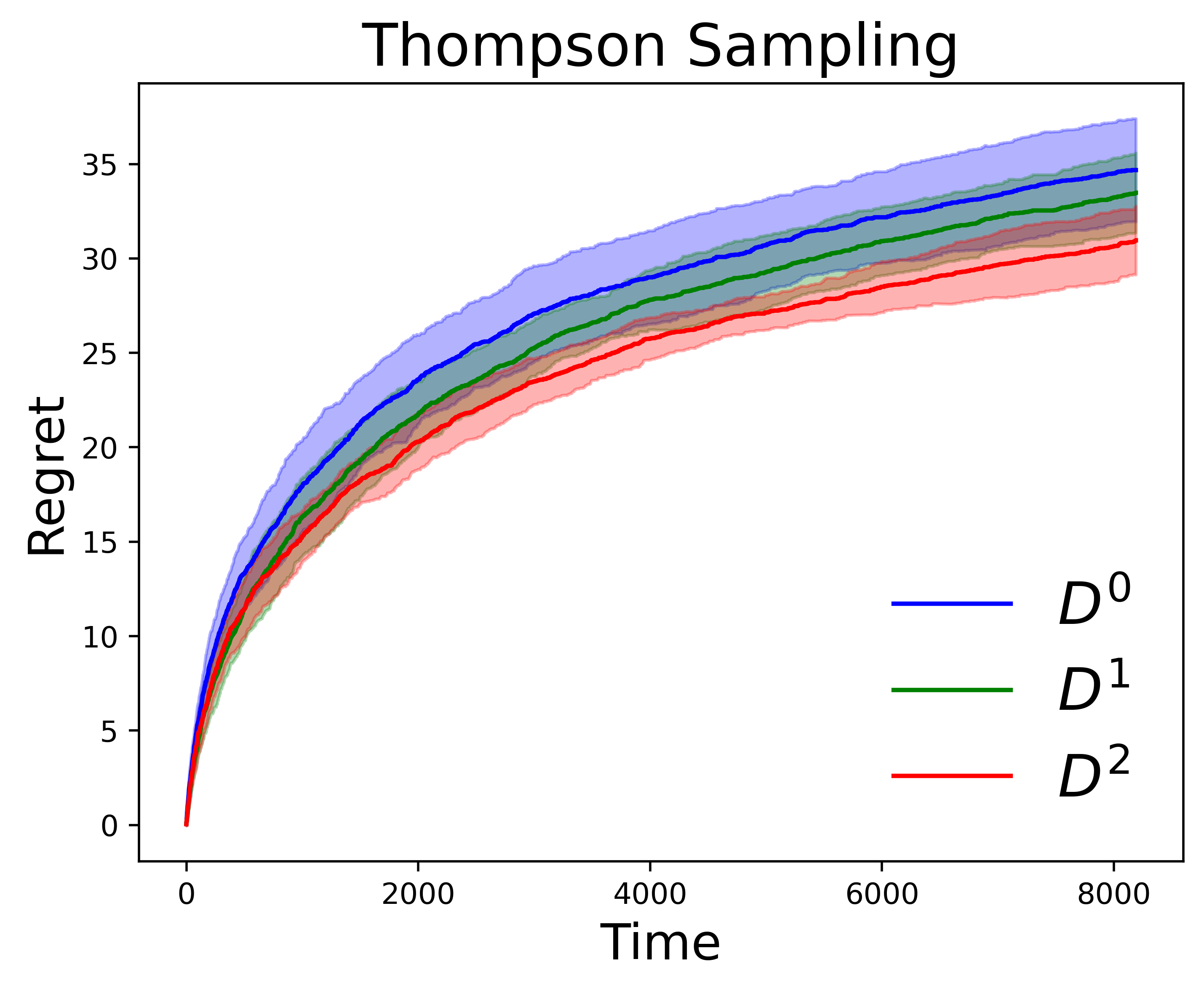}
        \caption{$\ts$ with extra info.} 
        \label{fig:ts-side-info}
    \end{subfigure}
    \begin{subfigure}{.33\linewidth}
        \centering
        \includegraphics[width=\linewidth]{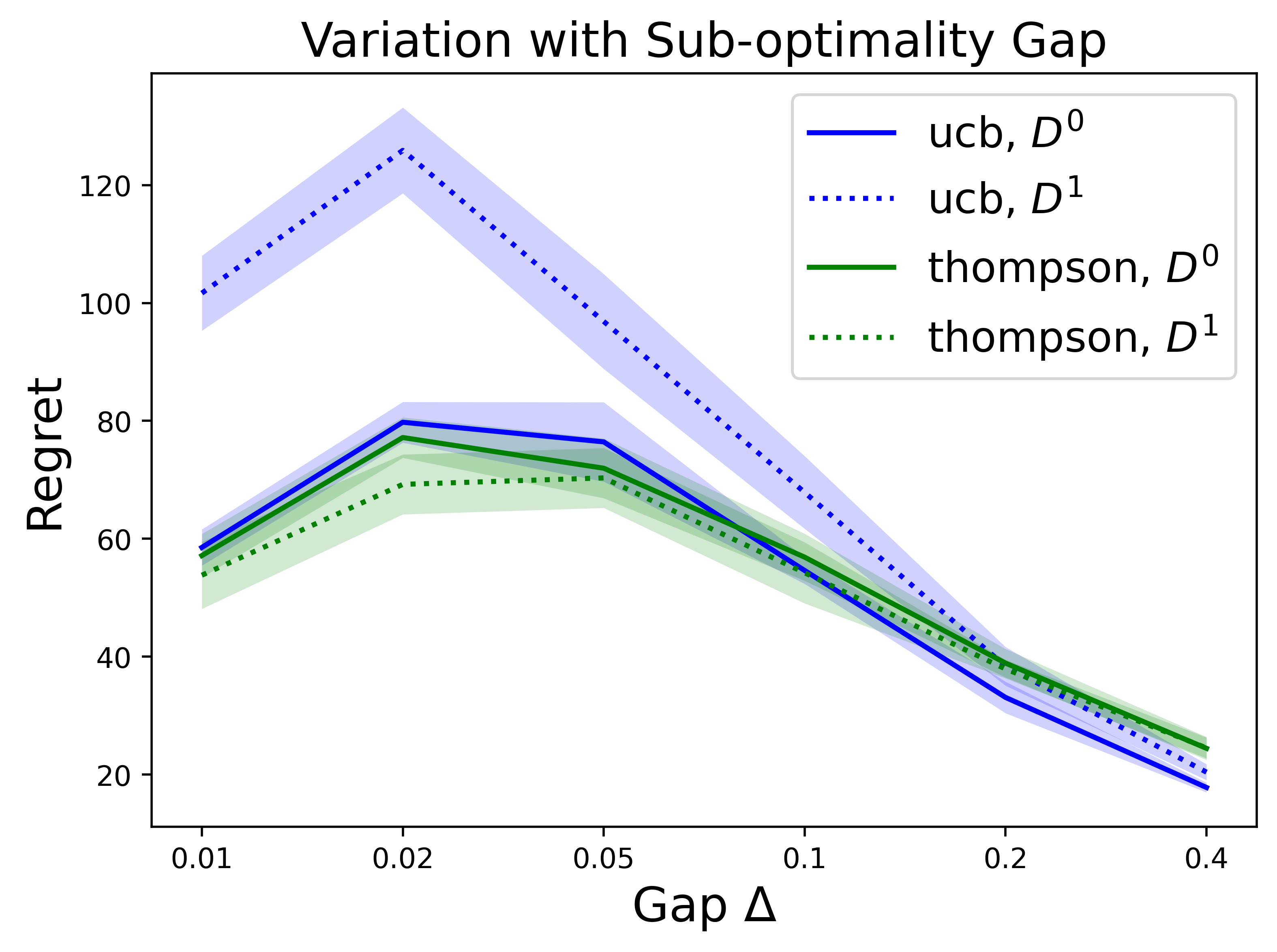}
        \caption{Effect of varying the gap $\Delta$.}
        \label{fig:vary-gaps}
    \end{subfigure}
    \caption{
    We compare the Upper Confidence Bound algorithm ($\ucb$) and the Thompson Sampling ($\ts$) algorithm on a two-armed bandit instance with action mean rewards $\mu=\left( \nicefrac{1}{2}+\Delta,\nicefrac{1}{2} \right)$ and $\Delta=0.1$ under three extra-information designs, where $\dd^i$ provides $i$ additional unbiased, independent samples of the optimal action at each time step.
    \cref{fig:ucb-side-info} shows increasing extra information consistently increases the regret of $\ucb$, while \cref{fig:ts-side-info} shows monotone regret reduction for Thompson Sampling. \cref{fig:vary-gaps} demonstrates that this qualitative contrast persists across different gaps $\Delta$, with larger regret increase for $\ucb$ on harder (smaller-gap) instances.
    }
    \label{fig:movielesn-occu-state}
\end{figure*}

We show that the widely used optimism-based Upper Confidence Bound ($\ucb$) algortithm \citep{auer_finite-time_2002} can react pathologically to extra information.
Extra samples---particularly that of `good' actions---can suppress optimism and
induce excessive exploration of inferior actions.
By contrast, algorithms such as Thompson Sampling \citep{thompson1933likelihood} appear to convert the same extra information into improved decisions and lower regret.
\cref{fig:movielesn-occu-state} illustrates this phenomenon on a simple
two-armed stochastic bandit:
providing increasing extra information about the optimal action consistently
\emph{increases} the regret of $\ucb$, while it \emph{decreases} the regret of
Thompson Sampling.

\subsection{Our Contributions}
We study the impact of exogenous extra information on stochastic bandit algorithms and identify a fundamental contrast between regret optimality and robustness to additional data. 
We formalize notions of non-degradation and monotonicity under extra
information (\cref{dfn:side-info-nondegrading,dfn:side-info-dominance,dfn:side-info-monotone}), that capture the desirable requirement that providing free and unbiased data should not increase the expected regret.
Our main contributions are:
\begin{itemize}
\item \textbf{$\ucb$ can degrade.}
We show that the canonical $\ucb$ algorithm violates non-degradation: there exist benign extra-information designs under which $\ucb$ incurs strictly higher regret
than with no extra information (\cref{thm:ucb-degrades}), and quantify this degradation (\cref{thm:ucb-zn-degradation}) in a simplified setup isolating the environment noise.
\item \textbf{Algorithms that benefit from extra  information.}
We discuss how extra information improves Thompson Sampling (\cref{thm:ucb-ts-difference}), and show that Explore-Then-Commit algorithm benefits from extra information (\cref{thm:etc-monotone}).
\item \textbf{Enforcing non-degradation.}
We propose a meta-algorithm that enforces regret `monotonicity' 
(i.e., lesser regret under more extra information) 
for a broad class of bandit algorithms (\cref{thm:meta-algo-monotone}).
\end{itemize}

\subsection{Related Work}

For general introductions to multi-armed bandits algorithms, see \citet{lattimore2020bandit,slivkins2019introduction}.
\rama{works on "how well side-info can be incorporated?" and how regret bounds can be lowered. 
Here, we only analyse algorithms and monotonicity property.}

\paragraph{Structured Extra Information.}
A line of work studies bandits in which the learner observes additional feedback beyond the reward of the played action.
Early formulations model this through \emph{feedback graphs}
\citep{mannor2011bandits,alon2015online,buccapatnam2014stochastic}, where playing an action reveals rewards of neighboring actions in the graph.
Regret bounds in this literature depend on structural properties of graphs such as independence or clique cover numbers.



Subsequent works generalize this model to allow extra information of varying
quality, often assuming a known feedback structure that can be exploited via
estimation or optimization techniques
\citep{atsidakou2022asymptotically,kocak2016online,wu2015online}.
Other extensions consider time-varying action sets \citep{cortes2019online}, switching costs \citep{arora2019bandits}, heavy-tailed rewards \citep{gou2023stochastic},
or unknown and evolving feedback structures 
\citep{cohen2016online,lykouris2020feedback}, thereby reducing the learner's ability to anticipate future extra information.
All of the above setups impose explicit structural assumptions on how extra information is generated.
In contrast, our framework models extra information through an
\emph{extra-information design}, specifying the number of additional unbiased samples provided to the learner and its timing.
The learner has no knowledge of, nor control over, this design.
This abstraction strictly generalizes feedback-graph models and allows extra
information to be adversarially timed and quantified while the actual extra-information samples remain stochastically benign.

Related works also consider adversarial or hybrid  reward settings
\citep{arora2019bandits,chen2021understanding,kong2022simultaneously}.
In contrast, we study i.i.d.\ rewards with unbiased extra information and show
that degradation can still occur.
Most feedback-graph models limit observation to one additional sample per action per round.
Our setup allows an arbitrary number of samples for any action at any round.

\paragraph{Warm-start and Offline data.}
Warm-start and offline-to-online bandits address the use of historical data to
mitigate `cold-start' effects, particularly in recommender systems
\citep{silva2023user,oetomo2023cutting,xie2023efficient}.
From our perspective, these settings can be viewed as special cases of extra-information designs in
which samples are front-loaded at initialization.

Existing approaches typically modify algorithms to explicitly incorporate
prior data, for example, by adjusting confidence bounds or priors
\citep{sharma2020warm,he2024learning}.
In contrast, we treat the learning algorithm as fixed and ask whether
additional unbiased samples will always be inherently beneficial.
Our results show that even in the simplest stochastic setting, this need not hold.
Related work also studies offline-to-online bandits under distribution shift
\citep{he2024learning}; in contrast, we focus on i.i.d.\ reward and
extra information samples, and show that regret increase can occur even in this benign setting.

\paragraph{Multi-agent and group-based learning.}
Prior works on multi-agent and group-based bandits often focus on minimizing
\emph{social regret}, i.e., the cumulative regret of all agents put together.
\citep{raghavan2018externalities,baek2021fair,krishnamurthy2026creator}.
These works show that $\ucb$-style algorithms can induce unfairness by forcing
some users (agents) to explore on behalf of others,
and focus on optimizing for fairness objectives under centralized control.
Our setting offerse a complementary decentralized perspective, where each user group can be interpreted as an independent learner running its own bandit algorithm and the statistics shared from other agents in the multi-agent collaboration are abstracted as free extra information given to the learner.
We show that even when observing extra information doesn't directly cause regret to the learner, it can modify the trajectory of $\ucb$ in a way that increases the regret.

On the other side of the spectrum in the multi-agent bandit learning scenarios are \emph{decentralized} models in which multiple agents privately interact with the environment over the full time horizon and share information among themselves at certain time steps to improve estimation, and each agent typically runs its own instance of a bandit algorithm.

A growing body of work studies decentralized bandits under communication
constraints or structured communication graphs
\citep{wang2019distributed,agarwal2022multi,zhu2021distributed}, as well as
extensions with richer assumptions, including joint reward models \citep{chang2022online},
heavy-tailed rewards \citep{dubey2020cooperative},
robustness to strategic agents \citep{krishnamurthy_collaborative_2026},
and robustness to Byzantine agents
\citep{zhu2024byzantine,dubey2020private}.
Across all these settings, multiple learning agents use variants/adaptations of the UCB paradigm to play actions and collaborate by sharing statitics.
Other algorithmic alternatives such as arm-elimination methods have also been explored \citep{zhangnear}.
Beyond multi-armed bandits, usage of optimism-based approaches is also prevalent in online and multi-agent reinforcement learning \citep{qiu2022contrastive,lee2022multi}.
In contrast to these decentralized multi-agent setups, our work can be interpreted as a per-agent perspective where statistics shared by other agents are received as extra information from the environment.

\section{Setting and Preliminaries}

Consider a learner facing a sequential decision-making problem
with a finite action set $\actions= \{1,2,\dots,K\}$ over a horizon of length $T$.
Playing an action (or arm) $i \in \actions$ yields an independent reward outcome from the environment (or nature)
drawn from an unknown sub-Gaussian distribution $r_i$ with unknown expected value $\mu_i \in [0,1]$.
At each time-step (or round) $t \in [T]$,
the learner plays an action $a(t) \in \actions$ and observes a reward $y(t) \sim r_{a(t)}$.
Denote this \emph{interaction \bu{h}istory} by
\begin{align}
H(t) = \left\{ \bigl( a(u), y(u) \bigr) : u < t \right\} \label{eqn:interaction-history}
\end{align}
to consist of all actions played and the corresponding rewards observed by the learner thus far up to the beginning of round $t$.
Let $\mathcal{H}$ denote the collection of all possible interaction histories $H(t)$s for all $t \in [T]$.

\paragraph{Extra information.}
At the beginning of each round $t \in [T]$, before playing any action, the learner may additionally be given some extra sample outcomes by the environment generated faithfully from the true distributions of the actions. 
The learner can use these extra samples to improve his action reward estimates.
This extra information is `free' in the sense that the learner does not incur any regret for observing these sample outcomes of the actions.
The quantity of this extra information is specified by a \bu{d}esign vector (or counts vector)
\[
\dd(t) = \bigl(\dd_1(t), \dd_2(t), \dots, \dd_K(t) \bigr) \in \nonnegint^K,
\]
where $\dd_i(t)$ denotes the number of outcome samples of action $i$ given by the environment at round $t$.
Concretely, at the beginning of round $t$, the environment reveals extra-infor\bu{m}ation set
\begin{align}
m_t = \left\{\left(i, \{y_{i,c}\}_{c=1}^{\dd_i(t)} \right) : i \in \actions \right\}, \label{eqn:set-of-side-info-samples-t}    
\end{align}
where for each action $i$, the samples $\{y_{i,c}\}_{c=1}^{\dd_i(t)}$ are all drawn from the true arm distribution $r_i$ independently across actions and rounds.

The full-horizon extra information design $\dd = \bigl(\dd(1),\dots,\dd(T) \bigr)$ is exogenous and fixed in advance: 
it does not depend on (and is not adaptive to) the learner’s actions, observations, or any internal randomness;
the learner cannot seek (or influence or control) which actions are sampled for extra information.
Further, this design $\dd$ is not known to the learner upfront, and is only noticeable in an online fashion when the actual extra-information samples $m_t$ are revealed at round $t$.
Let
$
M_t = \bigcup_{u \leq t} m_u
$
denote the \emph{extra-information history} consisting of all extra-information samples revealed so far to the learner upto the beginning of round $t$.
Let $\mathcal{M}$ denote the collection of all possible extra-information histories $M_t$ for all $t \in [T]$.

\paragraph{Algorithm and Regret.}
The learner employs an algorithm $\alg$ to select an action to play at each round $t \in [T]$,
based on the interaction history $H_t$ and the extra-information history $M_t$ available at the beginning of round $t$.
Formally, the algorithm is a (possibly randomized) mapping from the set of histories to the set of actions, i.e.,
$
\alg : \mathcal{H} \times \mathcal{M} \to \Delta(\actions),
$
where $\Delta(\actions)$ denotes the probability simplex over the action set.

The objective of the learner is to design an algorithm $\alg$ that minimizes the (expected) \bu{r}egret, defined as
\begin{align}
R(\alg, (\actions, \dd), T) 
\; = \;
\sum_{t=1}^{T} \mu_\star
\; - \;
\expund{\actions, \alg, \dd}{\mu_{a(t)}},
\label{eqn:regret}
\end{align}
where $\mu_\star = \max_{i \in \actions} \mu_i$ is the optimal expected reward.
The regret benchmarks the learner’s expected cumulative reward against that of
an oracle that always plays the optimal action.
We again note that observing extra-information samples does not make the learner incur regret, since they are observed without playing an action.

The expectation in \cref{eqn:regret} is taken over all sources of randomness:
(i) any internal randomness of the algorithm $\alg$,
and (ii) the stochasticity of the environment, 
which governs both the rewards observed from played actions 
and the samples revealed as extra information under the design $\dd$.


The action set and the extra-information design vector together specify the
sequential decision-making problem faced by the learner.
By a slight abuse of notation, we write $\actions = (r_1,\dots,r_K)$ to denote
the collection of reward distributions associated with the actions.
With that, we define the problem of \emph{bandit learning with extra information} by
the tuple $(\actions, \dd, T)$, with the learner’s goal being to minimize the regret in \cref{eqn:regret}.

When the design satisfies $\dd_i(t) = 0$ for all actions $i \in \actions$ and rounds $t \in [T]$
(denoted compactly as design $\dd~=~\boldzero$), the model reduces to the classical/vanilla multi-armed bandit (MAB) problem.
Thus, our setting strictly generalizes the standard MAB framework.

\paragraph{Non-degradation under extra information.}
A basic desideratum for learning algorithms---be it with offline or online training---is that giving additional data should not worsen performance.
In our setting of bandit learning with extra information,
it means consuming the free and unbiased extra-information samples should not increase the learner’s expected regret. 
We formalize this requirement as follows:
\begin{definition}[Non-degradation under extra information]
\label{dfn:side-info-nondegrading}
An algorithm $\alg$ is said to be \emph{non-degrading under extra information} if providing extra information never increases its expected regret.
Precisely,
for all bandit instances $\actions$, and all extra-information design vectors $\dd$, 
it holds that
\begin{align}
R(\alg, (\actions, \dd), T) 
\; \leq \;
R(\alg, (\actions, \boldzero), T),
\label{eqn:side-info-nondegrading}
\end{align}
for all horizons $T > T_0$ larger than some threshold $T_0$.
Here, $\boldzero = 0^{T \times K}$ denotes the design with no extra information.
\end{definition}
We simply say ``$\alg$ can degrade" if $\alg$ fails to satisfy \cref{dfn:side-info-nondegrading}.
We next state our first major result: The $\ucb$ algorithm can degrade.

\section{$\ucb$ can degrade with extra information}

A canonical example of a provably optimal stochastic bandit algorithm is
the $\ucb$ algorithm of \cite{auer_finite-time_2002}. 
It has become one of the most widely studied and deployed algorithmic paradigms for stochastic bandits.

At a high level, $\ucb$ augments empirical mean estimates with
optimism-based confidence bonuses to get a high-confidence upper \bu{b}ound on the the true reward mean of an action:
\iftwocolumn
\begin{align}
B_i(t) = \muhat_i + \sqrt{\tfrac{2 \log t}{n_i(t)}},
\end{align}
\else
$B_i(t) = \muhat_i + \sqrt{\nicefrac{2 \log t}{n_i(t)}}$,
\fi
where $n_i(t)$ is the total number of samples of action $i$ available. 
Then, it selects the action with the largest such upper confidence bound, $a_t \in \argmax_{i} B_i(t)$.
This idea gracefully interleaves and balances exploration and exploitation and guarantees sufficient
exploration of suboptimal actions in the absence of extra information.

Given this design, it is natural to expect $\ucb$ to benefit from extra-information samples.
Extra samples sharpen empirical estimates and facilitate faster identification of the optimal action.
However, this intuition overlooks a subtle but fundamental feature of $\ucb$:
the confidence bonus shrinks with the number of samples $n$,
scaling as $\simeq \sqrt{1/ n}$.
As a result, extra information about an action \emph{lowers} the action’s upper confidence
bound, thereby suppressing the optimism about that action.
When those extra-information samples come from the optimal action, its upper confidence bound is lowered.
This makes the $\ucb$ algorithm less likely to play this optimal action, and instead more likely to play the sub-optimal actions, thereby increasing the regret. 
We have
\begin{restatable}{theorem}{UcbDegrades}
\label{thm:ucb-degrades}
\upshape
$\ucb$ is \emph{not} non-degrading under extra information.
\end{restatable}
We defer the proof to \cref{appn-sec:ucb-degrades-proof} and give an outline here.
The result is established by constructing a problem instance where $\ucb$ degrades, i.e., incurs more regret with extra information than without it.
Consider a two-action stochastic bandit instance with action $1$ being uniquely optimal,
and an extra-information design $\dd$ that provides some $s>0$ samples of only the optimal action at the initial round, and no samples for either actions thereafter:
\begin{restatable}{equation}{UcbCounterExample}
\begin{aligned}
\dd(1) = (s, 0), \qquad
\dd(t) = (0, 0) \quad \text{for all } t \ge 2 .
\end{aligned}
\end{restatable}
The intuition is as follows.
By receiving samples of the optimal action upfront, $\ucb$ behaves as if it
had already explored this optimal action sufficiently.
This immediately shrinks the optimal action’s confidence interval (or bonus term) and lowers its upper confidence bound.
Consequently, $\ucb$ algorithm, which plays actions with the largest upper confidence bound, 
has less incentive to pull the optimal action and instead 
deterministically reallocates effort toward playing the suboptimal action in order to `catch up' and reduce its uncertainty about the sub-optimal action, each such pull incurs regret.

In our result, importantly, the side information is unbiased and actually informative: it consists entirely of samples drawn from the true reward distribution of the optimal
action, yet the regret increases.
The degradation arises solely from $\ucb$’s deterministic optimism-based
exploration strategy.
By front-loading information about the optimal action, the algorithm suppresses its own optimism for that action and over-explores the suboptimal actions, resulting in strictly higher regret.

\subsection{Zero-noise thought experiment}
The degradation of $\ucb$ is intrinsic to the algorithm, and not an artifact of reward noise.
To isolate the mechanism, we consider a thought experiment in which all external randomness is removed and rewards are deterministic, i.e., each action $i$ yields reward $\mu_i$ with probability one, and every extra-information sample for action $i$ is also exactly $\mu_i$.
We run $\ucb$ and Thompson Sampling faithfully, without informing them that the problem instance is noiseless.
In this setting, $\ucb$'s action-selection trajectory is completely deterministic and fully determined by the bandit instance and extra-information design,
whereas Thompson Sampling's trajectory remains stochastic due to its internal randomness in the posterior sampling step.
\paragraph{Effect on $\ucb$.} 
Empirical means of the actions equal their true means, so the upper confidence bound becomes
\begin{align*}
B_i(t) = \mu_i + \sqrt{\tfrac{2 \log t}{n_i(t)}},
\end{align*}
where $n_i(t)$ counts both interaction and extra-information samples. 
When extra information for action $i$ is given, $n_i(t)$ increases and consequently the bonus term decreases, thereby lowering $B_i(t)$. But, the upper confidence bounds of other actions remain unchanged.~\footnote{An alternative confidence term is $\sqrt{\tfrac{2 \log N_t}{n_t(i)}}$ where $N_t = \sum_i n_t(i)$ is the total samples over all actions. This choice also yields the same result of degradation.}
Thus, relateive to the no extra-information case, providing extra-information samples of the optimal action can only make the optimal action \emph{less} attractive to play for $\ucb$, and not more attractive.

For the optimal action, say, action $1$, as its upper confidence bound $B_1$ lowers,
the gap $B_1(\cdot) - B_j(\cdot)$ lowers for all sub-optimal actions $j \neq 1$, and $\ucb$ deterministically reallocates pulls to all these suboptimal actions to reduce their uncertainty, i.e., their upper confidence bounds $B_j(\cdot)$s. 
Thus, free extra-information samples of the optimal action lower its upper confidence bound, thus, increase the number of suboptimal pulls, and hence the regret.

\paragraph{Quantification of $\ucb$'s regret increase.}
Next, we try to quantify $\ucb$'s increase in regret as a function of the instance $\actions$ and extra-information design $\dd$. Towards that, we introduce some notation.
For a given $\dd$, write $s_i := \sum_{t=1}^T \dd_i(t)$ to be the total extra-information \bu{s}ample count of action $i$ 
over the course of the time horizon $T$.
We shall quantify the degradation lower bound as a function of these total counts $s_i$.

\begin{restatable}{theorem}{UcbZeroNoiseDegradation}
\label{thm:ucb-zn-degradation}
\upshape
Let $\actions$ be any deterministic bandit instance with $K$ actions, and assume w.l.o.g. that action $1$ is optimal.
For any extra-information design $\dd$, the degradation of $\ucb$ is lower bounded as follows:
\iftwocolumn
\begin{align*}
& R(\alg, (\actions, \dd), T) - R(\alg, (\actions, \boldzero), T)
\geq \\
& \sum_{i=2}^{K} \frac{2 (\alpha \ln T)^{1.5}}{\Delta_i^2} \left( \frac{1}{\sqrt{T}} -\frac{1}{\sqrt{T+s_1}} \right) - s_i \Delta_i - o(1). \numberthis \label{eqn:ucb-zn-degradation}
\end{align*}
\else
\begin{align*}
R(\alg, (\actions, \dd), T) - R(\alg, (\actions, \boldzero), T)
\geq \sum_{i=2}^{K} \frac{2 (\alpha \ln T)^{1.5}}{\Delta_i^2} \left( \frac{1}{\sqrt{T}} -\frac{1}{\sqrt{T+s_1}} \right) - s_i \Delta_i - o(1). \numberthis \label{eqn:ucb-zn-degradation}
\end{align*}
\fi
where $\Delta_i := \mu_{\star} - \mu_i$ is the sub-optimality gap of action $i \in \actions$.
\end{restatable}

\noindent The Theorem gives three interpretational take-aways.

First, the $s_1$ samples of optimal action act as a global spoiler. Because the upper confidence bound of the optimal action is a benchmark that every suboptimal action is compared against, lowering that benchmark unleashes unnecessary exploration across all $K-1$ suboptimal actions. Thus, the degradation scales linearly with the size of the action set as indicated by the summation.

Second, the degradation is proportional to $\nicefrac{1}{\Delta_i^2}$ which is worse than the standard $~\nicefrac{1}{\Delta_i}$ instance-dependent regret term \citep{auer_finite-time_2002,lai1985asymptotically}.
This indicates that for actions near optimal, 
providing extra-information of optimal action will cause more degradation, as the algorithm will play near-optimal actions more, as it is unable to use the optimal action's uncertainty (as the $s_1$ samples suppress it) to filter them out. This is also seen empirically in \cref{fig:vary-gaps} with higher degradation (i.e., the difference between the dotted and solid blue curves) observed for lower gaps $\Delta$.

Third, extra infomation of the sub-optimal actions gives a direct linear $s_i \Delta_i$ relief from the degradation, where the algorithm benefits from not having to explore by itself those sub-optimal actions. 

Aside, the $o(1)$ slack is independent of time horizon $T$ and extra-information sample counts $(s_i)_i$, and is merely an artifact of our analysis that uses Taylor approximations.

\paragraph{Effect on Thompson Sampling ($\ts$).} 
For ease of exposition, we consider bandit instances $\actions$ with gaussian reward distributions. 
So, $\ts$ maintains Gaussian prior and posterior distributions about the actions' true mean reward.
With deterministic rewards, empirical means equal true means, so the posterior distribution for action $i$ is 
\begin{align}
P_i(t) = \gaussian{\mu_i}{\tfrac{1}{n_i(t)}}.
\end{align} 
We recall that, at each round $t$, Thompson Sampling draws independent samples $\tilde{\mu}_i \sim P_i(t)$ from the current posterior, and plays an action $a_t \in \argmax_{i} \tilde{\mu}_i$ with the largest posterior sample.

For two actions $i,j$ such that $\mu_i > \mu_j$, the sampled difference $\tilde{\mu}_i(t) - \tilde{\mu}_j(t)$ is Gaussian with a positive mean $\mu_i - \mu_j$ and variance $\nicefrac{1}{n_i(t)} + \nicefrac{1}{n_j(t)}$ that depends inversely on the counts of the samples.
Observing extra-information samples about either action strictly reduces this variance while preserving the positive mean, thereby increasing $\prob{\tilde{\mu}_t(i) > \tilde{\mu}_t(j)}$, which is exactly the probability of selecting the more optimal action, thereby decreasing the regret. \\

In short, the same extra information suppresses $\ucb$'s optimism for the optimal action and increases the regret, whereas in $\ts$, it concentrates the posterior distributions around the true action reward means and reduces the regret.
We make this contrast formal: there is a minimal instance on which extra information has opposite effects on the two algorithms; the proof is in \cref{appn-sec:one-shot-comparison}.

\begin{restatable}{lemma}{UcbTsDifference}
\label{thm:ucb-ts-difference}
\upshape
There exists a stochastic instance $(\actions,\dd,T)$ of the bandit learning with extra information problem such that
\begin{align*}
R(\ucb, (\actions, \boldzero), T) & < R(\ucb, (\actions, \dd), T) \quad \text{ and } \quad \\
R(\ts, (\actions, \boldzero), T) & > R(\ts, (\actions, \dd), T).
\end{align*}
\end{restatable}
\section{Data-monotone algorithms}
We shall strengthen the notion of a `non-degrading' algorithm further and we shall show algorithms that satisfy that stronger property.
Towards that, we formalize the idea that
some extra-information designs are intrinsically more informative than others.
Intuitively, receiving additional samples of an action cannot reduce the information available to the learner, and receiving samples earlier in the horizon is at least as
informative as receiving them later.
Indeed, a learner that is given more samples, or receives them sooner, can always ignore some samples or postpone their use in order to emulate a less informative design.
This intuition induces a natural partial order over extra-information designs:
\begin{definition}[Dominance of extra-information designs]
\label{dfn:side-info-dominance}
A extra-information design $\ee$ is said to \emph{dominate} another design $\dd$, written $\ee \succeq \dd$,
if $\ee$ provides at least as many samples of every action as $\dd$ does by every point in time.
Formally, $\ee \succeq \dd$ if for all actions $i \in \actions$ and all time steps $t \in [T]$, it holds that
\begin{align}
\sum_{u=1}^{t} \dd_i(u) 
\; \leq \; 
\sum_{u=1}^{t} \ee_i(u). 
\label{eqn:side-info-dominance}
\end{align}
\end{definition}

This dominance relation captures both the quantity and the timing of extra-information samples.
We can therefore ask whether an algorithm’s performance improves monotonically as the available side information becomes more informative:
\begin{definition}[Data-monotone]
\label{dfn:side-info-monotone}
An algorithm $\alg$ is said to be \emph{data-monotone} if 
for any action set $\actions$ and any two extra-information designs $\ee \succeq \dd$,
\begin{align}
R(\alg, (\actions, \ee), T) 
\; \leq \;
R(\alg, (\actions, \dd), T),
\label{eqn:side-info-monotone}
\end{align}
for all time-horizons $T > T_0$ beyond a threshold $T_0$.
\end{definition}

Note that the design with no extra information,
$\boldzero~:=~(\dd_t(i) = 0 : \forall i \in \actions,\; t \in [T])$,
is dominated by every other extra-information design.
Consequently, non-degradation under extra information
(\cref{dfn:side-info-nondegrading}) is recovered as a special case of
data monotonicity
(\cref{dfn:side-info-monotone}).
In particular, any algorithm that is monotone under dominating extra information
is necessarily non-degrading under extra information.
We next show that there exist algorithms that satisfy this stronger property.

\subsection{Explore-Then-Commit is monotone}

We have seen in \cref{thm:ucb-ts-difference} that $\ucb$ can degrade under extra information, whereas Thompson Sampling improves in those same problem instances.

This raises a natural question:
Is internal randomization necessary for an algorithm to be non-degrading under extra information?
The answer is no.
In particular, simple deterministic algorithms can also satisfy monotonicity.

A canonical example is the Explore-Then-Commit ($\etc$) algorithm.
Recall that $\etc$ explores by pulling each action in a round-robin fashion for a fixed number of rounds $T^{ex} < T$, and then commits to the empirically best action for the remainder of the horizon.
With an appropriate choice of $T^{ex}$, it is known that this strategy achieves regret $\appxbigO{T^{\nicefrac{2}{3}}}$.
Next, we show that this algorithm is monotone:
\begin{restatable}{theorem}{EtcMonotone}
\label{thm:etc-monotone}
The $\etc$ algorithm is monotone under dominating side information.
\end{restatable}
The proof is deferred to \cref{appn-sec:etc-monotone-k}.
At a high level, the argument is straightforward:
providing dominating extra-information samples---earlier or in greater quantity---
can only reduce the variance of the corresponding empirical estimate.
Since $\etc$’s decision is made via a single comparison step at the end of the
exploration phase, reduced variance can only decrease the probability of
committing to any suboptimal action.

This behaviour contrasts sharply with $\ucb$.
For $\etc$, side information about the optimal action strictly increases the probability of committing to it.
For $\ucb$, the same information can instead suppress optimism for the optimal action and induce additional exploration of suboptimal ones.
This comparison highlights that degradation is not an inherent consequence of determinism, but of how exploration pressure is coupled to the observed data samples.



\section{Meta-Algorithm: Enforcing Non-Degradation}

\begin{algorithm}[t]
\small
\caption{$\nondegalg$: Deferred and balanced consumption of extra information}
\label{alg:deferred-balanced-learning-algo}
{\bf Input:} Black-box learning algorithm $\alg$, action set $\actions$.\\
{\bf Online input:} Extra-information samples set $s_t$ as in \cref{eqn:set-of-side-info-samples-t}.
\begin{algorithmic}[1]
\StatePar{For all actions $i \in \actions$, initialize buffers $S_i \gets \emptyset$ and counters $c_i \gets 0$} 
\label{ln:initialization}
\For{time-step $t = 1, 2, \dots, T$}
    \StatePar{Receive extra-information samples and put in buffers $S_i$}
    \label{ln:update-buffer}
    \StatePar{ Query the black-box algorithm for an action recommendation: $i \gets \alg(H_t)$}
    \label{ln:get-action-to-play-1}
    \While{$S_i \neq \emptyset$ \textbf{and} $c_i \le \min_{j \in \actions} c_j$}
        \label{ln:condition-for-side-info}
        \StatePar{$s \gets$ Remove one buffer sample from $S_i$}
        \StatePar{Append $(x=i,y=s)$ to history $H_t$}
        \StatePar{Increment buffer consumption counter $c_i$}
        \StatePar{Query the black-box algorithm again: $i \gets \alg(H_t)$}
        \label{ln:get-action-to-play-2}
    \EndWhile
    \StatePar{Play action $x_t=i$ on the environment and observe outcome $y_t$}
    \label{ln:play-action-on-env}
    \StatePar{Append $(x_t,y_t)$ to history $H_{t+1}$}
    \label{ln:action-outcome-consumed}
\EndFor
\end{algorithmic}
\end{algorithm}

The negative result in \cref{thm:ucb-degrades} shows that $\ucb$ algorithm naively incorporating extra information can suffer increased regret.
We now ask a more constructive question:
Can extra information be incorporated in a way that provably avoids degradation?
Our answer is affirmative.
We present a meta-algorithm that transforms any reasonable bandit algorithm into one that is monotone in dominating extra information, and thus is also non-degrading, for any arbitrary set of actions.

The key observation is that extra information differs from action play in that it provides samples without incurring regret;
however, if such samples are fed to a learning algorithm in an unbalanced manner, they may distort its exploration behaviour and lead to worse cumulative performance.
Our approach is therefore based on two principles: 
(1) Deferred consumption: extra information is not consumed immediately upon arrival;
(2) Balanced consumption: extra information is consumed evenly across actions.
These ideas shall ensure that extra information acts as `free exploration' that cannot result in more regret than exploring by playing that action.

Our approach works for all learning algorithms that satisfy: 
\begin{property}[Better-than-random algorithm]
\label{assm:alg-plays-subopt-fewer}
An algorithm $\alg$ is said to be \emph{better than random} if 
for any action set $\actions$,
there exists a threshold time $T'$ such that for all time horizons $T > T'$ and all window lengths $w \geq 1$,
the expected regret of $\alg$ over the window $[T : T+w]$ is upper bounded by the regret of playing all actions uniformly at random over the same window.  
Precisely,
\begin{align*}
R(\alg, (\actions, \boldzero), [T : T + w]) \leq \frac{w}{K} \sum_{i \in \actions} \Delta_i,
\end{align*}
where $K=\cardinality{\actions}$ and $\Delta_i = \mu_* - \mu_i$ denotes the sub-optimality gap of action $i$.
\end{property}
\noindent This property is mild and asks that an algorithm learn enough by some suitably large time-step $T$ so as to thereafter perform better than a policy that plays/explores all actions uniformly at random (or in round robin).
It is easily seen that this property is satisfied by standard bandit algorithms such as $\ucb$, Thompson Sampling etc. 
which all have known asymptotic sub-linear regret upper bounds, whereas performing uniform exploration at all times would incur a linear regret.

\noindent \cref{alg:deferred-balanced-learning-algo} presents the meta-algorithm $\nondegalg$, which controls the manner in which extra information is consumed by a black-box learning algorithm $\alg$.


\paragraph{Meta-algorithm description.} 
The algorithm maintains a buffer of unused extra-information samples for each action, as well as counters that track the number of extra information samples consumed per action (line~\ref{ln:initialization}).
At each round $t$, the extra information samples revealed at that time are first stored in their corresponding buffers and are not immediately consumed (line~\ref{ln:update-buffer}).
Using the current history $H_t$, the meta-algorithm then queries the black-box learner $\alg$ to obtain an action recommendation (line~\ref{ln:get-action-to-play-1}).
If an extra-information sample for the recommended action is available in the buffer, and the total number of previously consumed extra information samples for that action does not exceed the minimum consumption across all actions, the meta-algorithm consumes one such sample. (line~\ref{ln:condition-for-side-info}).

The sample is appended to the history without interacting with the environment, and the black-box algorithm is queried again (line~\ref{ln:get-action-to-play-2}).
This procedure repeats until either no extra information is available for the recommended action or the balanced-consumption condition is violated.
Once no further extra information can be consumed, the meta-algorithm executes the recommended action in the environment, observes the resulting outcome, (incurs regret,) and appends it to the history for the next time-step (line~\ref{ln:action-outcome-consumed}).
This process continues for the entire time horizon.
Deferring the proof to \cref{appn-sec:meta-algo-proof} in the interest of space, we have



\begin{restatable}{theorem}{MetaAlgoMonotone}(Enforcing data-monotonicity.)
\label{thm:meta-algo-monotone}
\upshape
When run with any learning algorithm $\alg$ that is better than random (\cref{assm:alg-plays-subopt-fewer}),
$\nondegalg$ is monotone in dominating extra information.
\end{restatable}

While our $\nondegalg$ algorithm ensures that regret weakly decreases when extra information increases, a natural question arises: Does the $\nondegalg$ algorithm make optimal use of the extra information?
Does it achieve the maximum regret reduction possible? 

First, our $\nondegalg$ is a wrapper algorithm that intrinsically depends on the decision-making of an internal bandit algorithm $\alg$ that can degrade. 
So, it is unlikely to achieve the maximum regret reduction, and it is to be stressed that the design goal of $\nondegalg$ is towards provably making $\alg$ robust to degradation.

Second, quantifying the maximum possible regret reduction with extra information by any possible bandit algorithm is a fundamental information-theoretic question.
It is possible that the regret minimality for different structures of extra-information designs are obtained by different bandit algorithms which take advantage of specific structural aspects. 
We leave this as an interesting direction for future work.

\subsection{Ablation study of $\nondegalg$.}

\begin{figure*}[ht]
\centering
\includegraphics[width=0.99 \linewidth]{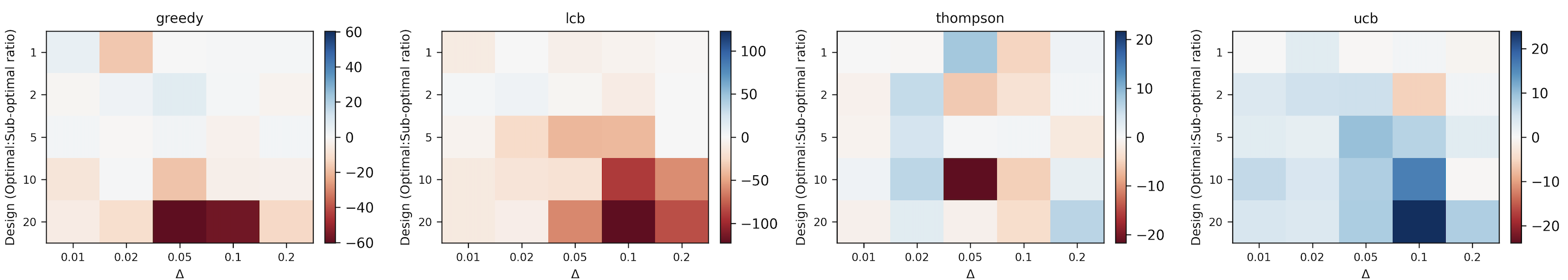}
\caption{An ablation study of how regret with extra-information is provided, across different algorithms in the different plots. 
In each, we plot the regret difference between the normal algorithm and that of wrapped in $\nondegalg$, plotted across different gaps $\Delta$ and different ratios of optimal-to-suboptimal extra-information samples.}
\label{fig:ablation}
\end{figure*}

Denote by $\nondegalg + \alg$ the execution of our $\nondegalg$ algorithm with an internal black-box bandit algorithm $\alg$. 
Then, for any two extra-information designs $\ee \succeq \dd$ (i.e., $\ee$ dominates $\dd$ as in \cref{dfn:side-info-dominance}) 
and for any bandit problem instance $I$, for a suitably large $T$,
our \cref{thm:meta-algo-monotone} shows that 
\iftwocolumn
\begin{align*}
R(\nondegalg + \alg, (I, \ee), T) \leq R(\nondegalg + \alg, (I, \dd), T).    
\end{align*}
\else
$R(\nondegalg + \alg, (I, \ee), T) \leq R(\nondegalg + \alg, (I, \dd), T)$.
\fi

However, there is complementary question that arises: For a specific bandit instance $I$, and a fixed extra-information design $\dd$,
how is the regret of the learner affected by using the $\nondegalg$ algorithm on top the original bandit algorithm $\alg$?
Specifically, how does
\iftwocolumn
\begin{align*}
R(\alg, (I, \dd), T) - R(\nondegalg + \alg, (I, \dd), T) 
\end{align*}
\else
$R(\alg, (I, \dd), T) - R(\nondegalg + \alg, (I, \dd), T)$
\fi 
behave for different $I, \dd, \alg$ combinations?

We plot in \cref{fig:ablation} an ablation study across these parameters: one plot for each $\alg$, and different extra-information designs across the vertical axis, and different sub-optimality gaps $\Delta$ across the horizontal axis.
These plots help understand how the $\nondegalg$ algorithm affects the instance-wise regret, while the non-degradation/monotonicity property is guaranteed over dominating extra-information designs.

The $\nondegalg$ algorithm lowers $\ucb$'s regret across 24 out of the 25 cases. The lower rows which correspond to disproportionately high amount of optimal action samples causes the $\ucb$ to degrade more, and thus the benefit of using $\nondegalg$ algorithm is larger, as indicated by the more dark blue-ish mosaic.

For Thompson Sampling, for roughly half the cases, the difference is negative (orange in colour)
because Thompson Sampling doesn't inherently degrade with more extra information.
So, in those cases, the $\nondegalg$ algorithm unnecessarily delays the consumption of extra information and increases the regret.
A take-away from these plots is that the $\nondegalg$ algorithm is more suitable for algorithms like $\ucb$ that are known to degrade.

For algorithms with no exploration (like Greedy and the Lower Confidence Bound (LCB) algorithms), our meta-algorithm behaves erratically (dark mosaics).
We do not expect them to perform well as both these algorithms are not necessarily better than random, i.e., they violate \cref{assm:alg-plays-subopt-fewer}.

\section{Conclusion}
We show that the classical intuition that more data is always beneficial can fail in sequential decision-making.
By formalizing notions of non-degradation and monotonicity to dominating side information, 
we demonstrate that $\ucb$---despite its optimal regret guarantees in the vanilla setup---can incur higher regret when provided with benign additional data.
In contrast, Thompson Sampling and Explore-Then-Commit benefit from side-information in similar instances.

We show that the effect of side information depends on how algorithms translate information into exploration.
We also propose a meta-algorithm that can ensure that any reasonable bandit algorithms do not suffer regret increase as more extra information is given to it.
In conclusion, 
our findings suggest a cautionary principle for incorporating extra information: \emph{More is \textbf{not} always merrier.}

\bibliography{references}
\bibliographystyle{plainnat}


\appendix

\section{Proof of \cref{thm:ucb-degrades}}
\label{appn-sec:ucb-degrades-proof}
\UcbDegrades*

\begin{proof}
We exhibit an online bandit problem with side information for which $\ucb$
violates \cref{eqn:side-info-nondegrading}.

Let $\actions = \{1,2\}$ be a two-action problem, where rewards are drawn from
distributions $r(1)$ and $r(2)$ with means $\mu(1) > \mu(2)$, so that action~$1$ is optimal.

Consider a side-information design $\dd$ that provides $s \geq 1$ samples of the optimal action at the initial round and no further side information thereafter:
\begin{align}
\dd_1=(s,0), \qquad 
\dd_t=(0,0)\ \quad \text{for all } t \geq 2. \label{eqn:counter-example-design-2}
\end{align}


To establish the theorem, it suffices to show that $\ucb$ incurs strictly larger expected regret under this design than in the absence of side information.
This is formalized in the following lemma. 
\begin{lemma}
\label{lem:ucb-degrades}
For the action set $\actions$ and side-information design $\dd$ defined in
\cref{eqn:counter-example-design-2}, it holds that
\[
R(\ucb, (\actions, \dd), T)
\; > \;
R(\ucb, (\actions, \boldzero), T)
\]
for all $T > 2$.
\end{lemma}
\begin{proof}
To formalize the argument, we adopt a coupling-based view of the environment.
Fix an infinite outcome tape (or film) $F = (F(1),F(2))$, which specifies, 
for each action $i \in \actions$, an infinite sequence of realized rewards
$F(i) = (f_1(i), f_2(i), \dots)$ drawn i.i.d.\ from that action's reward outcome distribution $r(i)$.

A run of $\ucb$---with or without side information---can then be simulated
deterministically from the tape $F$ 
by consuming the next unused tape entry whenever the algorithm either
(i) plays an action and receives a reward, or 
(ii) receives a side-information sample for that action.

Under this construction, all environmental randomness that manifest in the form of stochastic rewards and side-information samples is realized in advance by the tape $F$.
Since $\ucb$ itself is deterministic, the entire trajectory of the algorithm is uniquely determined by the tape $F$, the side-information design $D$, and the horizon $T$.

We define the \emph{tape regret} (or film regret)
$R^F(\ucb, (\actions, \dd), T)$ as the pseudo-regret incurred by running $\ucb$ for $T$ rounds on instance $(\actions, \dd)$ with action outcomes generated according to the tape $F$.
In particular, for the two-action instance considered here, the tape regret
is a nonnegative integer equal to the number of times the suboptimal action is played during this deterministic simulation.

The proof proceeds in two conceptual steps.
First, in \cref{clm:tape-regret-inequality}, we establish a \emph{pointwise} comparison : for every fixed outcome tape $F$,
the regret incurred by $\ucb$ without side information is at most the regret incurred with side-information design $D$.
Second, in \cref{clm:tape-regret-strict-inequality},  we show that this comparison is \emph{strict} on a set of tapes with strictly positive probability.
On these tapes, the early side-information samples of the optimal action are unlucky enough to permanently suppress its UCB index, causing $\ucb$ to play
the suboptimal action for the entire horizon under design $D$, while the
no-side-information run eventually plays the optimal action.
Together, these two claims imply that expected regret is strictly larger
under $D$.

We show that for every fixed tape $F$, the regret incurred by
$\ucb$ without side information is no larger than the regret incurred when
side information is provided according to the design $D$.
This pointwise comparison is captured by the following claim:
\begin{claim}[Tape regret inequality]
\label{clm:tape-regret-inequality}
For every outcome tape $F$,
\[
R^F(\ucb, (\actions, \boldzero), T)
\;\leq\;
R^F(\ucb, (\actions, \dd), T).
\]
\end{claim}
\begin{proof}
In this simulation, the $s$ side-information samples of the optimal action---namely $f_1(1), f_2(1), \ldots, f_s(1)$---are consumed from the tape at the very beginning of the simulation.
As a result, by the end of $T$ decision rounds, a total of $T+s$ samples have been
consumed from the two tapes $F(1)$ and $F(2)$ combined.

Let $n^{\textsc{D}}_2 \leq T$ denote the total number of samples consumed from the
suboptimal action’s tape $F(2)$.
Because the design $D$ provides no side information for action~$2$, each such
sample must correspond to an actual pull of the suboptimal action.
Hence, the tape regret under design $D$ is
\begin{align}
R^F(\ucb, (\actions, \dd), T) = n^{\textsc{D}}_2.    
\end{align}

We now compare this with a run of $\ucb$ for $T$ decision rounds with no
side information (design $\boldzero$), using the same tape realization $F$.
We show that, in this run, the suboptimal action can be played at most
$n^{\textsc{D}}_2$ times.

We distinguish two cases based on the trajectory induced by the run on $(\actions, \dd)$ over $T$ rounds.

\paragraph{Case 1: The optimal action is never played.}  
In this case, the suboptimal action is played in every round, so $n^{\textsc{D}}_2 = T$.
Trivially, 
\begin{align}
R^F(\ucb, (\actions, \boldzero), T) \le T = n^{\textsc{D}}_2.
\end{align}

\paragraph{Case 2: The optimal action is played at least once.}
Let $n^{\textsc{D}}_1 = T - n^{\textsc{D}}_2 > 0$ denote the number of times the optimal action is played under design $D$. 
In this case, the run of $\ucb$ has effectively progressed beyond the initial side-information phase, after which the evolution of the algorithm proceeds in a standard manner.


We argue by the desired result contradiction.
Suppose that, when $\ucb$ is run with no side information on the same tape $F$,
the suboptimal action is played for a $(n^{\textsc{D}}_2 + 1)$-th time at some
round $t' \leq T$.

Under design $D$, the run consumes exactly 
$s + n^{\textsc{D}}_1 = s + T - n^{\textsc{D}}_2$ samples from the optimal action’s tape $F(1)$, 
and $n^{\textsc{D}}_2$ samples from the suboptimal action’s tape $F(2)$.

Since $\ucb$ is deterministic given its observations,
we have in the run with design $\boldzero$ as well that
before the $n^{\textsc{D}}_2+1$-th sample from $F(2)$ is consumed, to match the same state, $s + T - n^{\textsc{D}}_2$ samples from $F(1)$ must have been consumed.

However, in the absence of side-information, all those samples are consumed exclusively via actual action pulls.
Therefore, to consume $n^{\textsc{D}}_2+1$ samples from $F(2)$ and 
$s+T-n^{\textsc{D}}_2$ samples from $F(1)$ via action pulls alone,
the algorithm must have run for at least 
$(n^{\textsc{D}}_2+1) + (s + T - n^{\textsc{D}}_2) = s + T + 1$ rounds.
This contradicts the assumption that $t' \leq T$.

We conclude that, in the run without side information, the suboptimal action is
played at most $n^{\textsc{D}}_2$ times. \\

\noindent Consequently, from both the cases above,
we have $R^F(\ucb, (\actions,0), T) = n^{\textsc{D}}_2 \leq R^F(\ucb, (\actions, \dd), T)$, 
completing the proof of the Claim.
\end{proof}

\noindent It remains to show that the regret inequality is strict on a set of tapes with strictly positive probability.

Fix $m := \sqrt{\frac{2 \log(T+s)}{s}}$, and define the set of \emph{unlucky tapes} 
\begin{align}
\mathcal{F}^{\varepsilon} := \left\{ F \in \mathcal{F} : \left( \frac{1}{s} \sum_{u=1}^{s} f_u(1) \leq \mu(2) - m \right) \land \Biggl( \forall t \in [T], f_t(2) > \mu(2) \Biggr) \right\}.  \label{eqn:unlucky-tape}
\end{align}
That is, the average of the first $s$ samples of the optimal action are unusually pessimistic, while all the samples of the sub-optimal action exceed its expected value.
For gaussian reward distributions, $\mathcal{F}^{\varepsilon}$ has strictly positive probability, $\probunder{\actions}{\mathcal{F}^{\varepsilon}} > 0$.

\begin{claim}
\label{clm:tape-regret-strict-inequality}
For any unlucky tape $F \in \mathcal{F}^{\varepsilon}$,
$R^F(\ucb, (\actions,\boldzero), T) < R^F(\ucb, (\actions,\dd), T) $.
\end{claim}
\begin{proof}
We first show that, for any unlucky tape $F \in \mathcal{F}^{\varepsilon}$, the run of $\ucb$ with side-information design $D$ never selects/plays the optimal action over the horizon $T$.

\noindent We show this by induction over time-step $t \in [T]$.

\textbf{Hypothesis. } Let $H(t)$ denote the event that action~$1$ has not been selected in the first $t$ decision rounds.
We prove that $H(t)$ holds for all $t\in\{0,1,\dots,T\}$.

\textbf{Base Case ($t=1$). }
After consuming the $s$ side-information samples of action~$1$ from the tape $F$, we have empirical mean $\muhat_{t=0}(1) \leq \mu(2) - m$ and and its upper confidence bound $\mathrm{UCB}_{t=0}(1) = \mu(2) - m + \sqrt{\frac{2 \log(0+s)}{s}} \leq \mu(2)$.

Action~$2$ has not yet been sampled, so its upper confidence bound index is strictly larger, and $\ucb$ selects action~$2$ at the first round.
Hence $H(1)$ holds.

\textbf{Inductive step. }
Assume $H(t)$ is true for some $t < T$, i.e., action~$1$ has not been selected in the first $t$ rounds.
Then the empirical mean and sample count of action~$1$ remain unchanged:
$\muhat_{t}(1) = \muhat_{0}(1) \leq \mu(2) - m$ and its upper confidence bound is 
\begin{align}
\mathrm{UCB}_{t}(1) = \mu(2) - m + \sqrt{\frac{2 \log(t+s)}{s}} \leq \mu(2), \label{eqn:unlucky-ucb-1}
\end{align}
where the last inequality follows from $t < T$ and the definition of $m$.

By contrast, action~$2$ has been selected in each of the first $t$ rounds.
Since $f_u(2) > \mu(2)$ for all $u$ on tapes in $\mathcal{F}^{\varepsilon}$, we have
\begin{align}
\mathrm{UCB}_{t}(2) = \muhat_t(2) + \sqrt{\frac{2 \log(t+s)}{t}} \labelrel{>}{i:8a} \mu(2), \label{eqn:unlucky-ucb-2}
\end{align}
where \eqref{i:8a} is due to the tape values obeying $f_u(2) > \mu(2)$ for all $u \in [T]$ as in \cref{eqn:unlucky-tape}.

Combining \cref{eqn:unlucky-ucb-1,eqn:unlucky-ucb-2}, we conclude that
$\mathrm{UCB}_t(2)>\mathrm{UCB}_t(1)$, so action~$2$ is selected at round $t+1$.
Thus $H(t+1)$ holds.

By mathematical induction, action~$1$ is never selected over the horizon $T$, and hence $R^F(\ucb, (\actions, \dd), T) = T$.

\paragraph{No-side-information run.}
In contrast, when $\ucb$ is run without side-information, the algorithm must sample each action at least once to initialize its estimates.
Therefore, action~$1$ is selected at least once, implying
$R^F(\ucb,(\actions,\boldzero),T) < T$. \\

\noindent Combining the two conclusions yields the claim.
\end{proof}

Taking expectation over the random tape realization $F$,
drawn according to the product distribution, say $P_\actions$, induced by the reward distributions $r(1)$ and $r(2)$, we obtain
\begin{align}
    R(\ucb, (\actions, \dd), T) = \expectation{R^F(\ucb, (\actions, \dd), T)} \labelrel{>}{i:8b} \expectation{R^F(\ucb, (\actions, \boldzero), T)} = R(\ucb, (\actions, \boldzero), T), \label{eqn:ucb-exp-reg-inequality}
\end{align}
where \eqref{i:8b} follows from the tape-wise inequality in \cref{clm:tape-regret-inequality} together with the strict inequality on a set of tapes of strictly positive probability established in \cref{clm:tape-regret-strict-inequality}.
Finally, \cref{eqn:ucb-exp-reg-inequality} shows the Lemma.
\end{proof}
This completes the proof of the theorem.
While the argument is demonstrated using a specific counter-example of a two-action instance and a particular side-information design $D$, 
similar phenomena can arise in settings with larger action sets or with alternative designs in which side-information is revealed gradually over time.
\end{proof}

\section{Proof of \cref{thm:ucb-zn-degradation}}

For a given problem instance $(\actions, \dd, T)$,
recall that the $\ucb$ algorithm maintains at time $t \in [T]$ an upper confidence \textbf{b}ound/index/envelope) $B^{\dd}_{i}(t)$ for each action $i \in \actions$ and plays the action $a(t) = \argmax_{i \in \actions} B^{\dd}_{i}(t)$.

Here, the upper confidence bound is given by $B^{\dd}_{i}(t) := \hat{\mu_{i}}(t) + \sqrt{\frac{\alpha \ln T}{n^{\dd}_{i}(t) + s^{\dd}_{i}(t)}}$,
where $n^{\dd}_{i}(t)$ is the 
number of times $\ucb$ plays action $i$ upto time $t$ when side-information is provided as per design $\dd$,
and $s^{\dd}_{i}(t)$ is the number of side-information samples of action $i$ provided upto time $t$ as per $\dd$.
When there is no side-information, i.e., design is $\boldzero$, we have $s^{\boldzero}_{i}(t) = 0$ for all $t,i$.

\label{appn-sec:ucb-zn-proof}
\UcbZeroNoiseDegradation*

\begin{proof}
To quantify the degradation of $\ucb$ because of side-information $\dd$ and show the inequality in \cref{eqn:ucb-zn-degradation},
we need to upper bound the regret with side-information $\dd$ and lower bound the regret when there is no side-information.
For this, we lower bound the number of times a sub-optimal action is played with $\dd$ in \cref{clm:dd-subopt-lower}, and upper bound the number of times a sub-optimal action is played with no side-information $\boldzero$ in \cref{clm:zero-subopt-upper}.
Finally, we use these to show \cref{eqn:ucb-zn-degradation}. 

\begin{claim}[Lower bound of sub-optimal action play in $\dd$]
\label{clm:dd-subopt-lower}
When $\ucb$ is run on $(\actions, \dd, T)$ for a suitable $T$, the number of times a sub-optimal action $i$ is played is lower bounded as follows:
\begin{align}
    n^{\dd}_{i}(T) \geq \frac{\alpha \ln T}{\Delta_i^2} \left( 1 - \frac{2}{\Delta_i} \sqrt{\frac{\alpha \ln T}{s_1 + n^{\dd}_{1}(T)}} \right) - s_i.
\end{align}
\end{claim}
\begin{proof}
Let $T$ be a time-step in which optimal action $1$ is played.
Then, its $UCB$ is higher than that of all the other actions. Thus, for any sub-optimal action $i$, we have 
\begin{align*}
   && \mu_i + \sqrt{\frac{\alpha \ln T}{s_i + n^{\dd}_{i}(T)}} & \;\leq\;  \mu_1 + \sqrt{\frac{\alpha \ln T}{s_1 + n^{\dd}_{1}(T)}} &\\
   \iff && \frac{\alpha \ln T}{s_i + n^{\dd}_{i}(T)} & \;\leq\; \left( \Delta_i + \sqrt{\frac{\alpha \ln T}{s_1 + n^{\dd}_{1}(T)}} \right)^2 &\\
   \iff && n^{\dd}_{i}(T) & \;\geq\; \frac{\alpha \ln T}{ \left( \Delta_i + \sqrt{\frac{\alpha \ln T}{s_1 + n^{\dd}_{1}(T)}} \right)^2} - s_i & \numberthis \label{eqn:dd-subopt-lower-1}
\end{align*}
Further simplifying \cref{eqn:dd-subopt-lower-1}, we have
\begin{align*}
n^{\dd}_{i}(T) & \geq   \frac{\alpha \ln T}{\Delta_i^2 \left( 1 + \frac{1}{\Delta_i} \sqrt{\frac{\alpha \ln T}{s_1 + n^{\dd}_{1}(T)}} \right)^2} - s_i \\
& \labelrel{\geq}{i:7-a} \frac{\alpha \ln T}{\Delta_i^2} \left( 1 - \frac{2}{\Delta_i} \sqrt{\frac{\alpha \ln T}{s_1 + n^{\dd}_{1}(T)}} \right) - s_i, 
\end{align*}
where \eqref{i:7-a} uses a first-order Taylor approximation $(1+x)^{-2} \geq 1 - 2x$ for $x>0$ with $x = \frac{1}{\Delta_i} \sqrt{\frac{\alpha \ln T}{s_1 + n^{\dd}_{1}(T)}}$.
\end{proof}

\begin{claim}[Upper bound of sub-optimal action play in $\boldzero$]
\label{clm:zero-subopt-upper}
When $\ucb$ is run on $(\actions, \boldzero, T)$, i.e., without any side-information,
the number of times a sub-optimal action $i$ is played is upper bounded as 
\begin{align}
    n^{\boldzero}_{i}(T) \leq \frac{\alpha \ln T}{\Delta_i^2} \left( 1 - \frac{2}{\Delta_i} \sqrt{\frac{\alpha \ln T}{n^{\dd}_{1}(T)}} + \frac{3}{\Delta_i^2} \frac{\alpha \ln T}{n^{\boldzero}_{1}(T)} \right).
\end{align}
\end{claim}
\begin{proof}
Let $T_i$ be the last time-step at which the sub-optimal action $i$ is played.
Then, its upper confidence bound is higher than that of the optimal action $1$. 
Thus, for any sub-optimal action $i$, we have
\begin{align*}
&& \mu_i + \sqrt{\frac{\alpha \ln T}{n^{\boldzero}_{i}(T_i)}} & \;\geq\;  \mu_1 + \sqrt{\frac{\alpha \ln T}{n^{\boldzero}_{1}(T_i)}} &\\
\iff && \frac{\alpha \ln T}{n^{\boldzero}_{i}(T_i)} & \;\geq\; \left( \Delta_i + \sqrt{\frac{\alpha \ln T}{n^{\boldzero}_{1}(T_i)}} \right)^2 &\\
\iff && n^{\boldzero}_{i}(T_i) & \;\leq\; \frac{\alpha \ln T}{ \left( \Delta_i + \sqrt{\frac{\alpha \ln T}{n^{\boldzero}_{1}(T_i)}} \right)^2} &  \\
\labelrel{\implies}{i:7-b} &&  n^{\boldzero}_{i}(T) & \;\leq\; \frac{\alpha \ln T}{ \left( \Delta_i + \sqrt{\frac{\alpha \ln T}{n^{\boldzero}_{1}(T)}} \right)^2}, \numberthis \label{eqn:zero-subopt-upper-1}
\end{align*}
where \eqref{i:7-b} uses the facts $n^{\boldzero}_{i}(T) = n^{\boldzero}_{i}(T_i)$ which holds by definition of $T_i$, and $n^{\boldzero}_{1}(T_i) \leq n^{\boldzero}_{1}(T)$ which holds by monotonicity of the play count.
Further simplifying \cref{eqn:zero-subopt-upper-1}, we have
\begin{align*}
n^{\dd}_{i}(T) & \;\leq\; \frac{\alpha \ln T}{\Delta_i^2 \left( 1 + \frac{1}{\Delta_i} \sqrt{\frac{\alpha \ln T}{n^{\boldzero}_{1}(T)}} \right)^2} \\
& \;\labelrel{\leq}{i:7-bb}\; \frac{\alpha \ln T}{\Delta_i^2} \left( 1 - \frac{2}{\Delta_i} \sqrt{\frac{\alpha \ln T}{n^{\boldzero}_{1}(T)}} + \frac{3}{\Delta_i^2} \frac{\alpha \ln T}{n^{\boldzero}_{1}(T)} \right), 
\end{align*}
where \eqref{i:7-bb} uses a second-order Taylor approximation $(1+x)^{-2} \leq 1 - 2x + 3x^2$ for $x>0$ with $x = \frac{1}{\Delta_i} \sqrt{\frac{\alpha \ln T}{n^{\boldzero}_{1}(T)}}$.
\end{proof}

Decomposing the regret of the algorithm as the sum of the number of times each sub-optimal action is played, and using its lower and upper bounds in \cref{clm:dd-subopt-lower,clm:zero-subopt-upper}, we quantify the degradation of $\ucb$ as
\begin{align*}
& R(\ucb, (\actions, \dd), T) - R(\ucb, (\actions, \boldzero), T) \;=\; \sum_{i=2}^{K} \Delta_i \times \left( n^{\dd}_{i}(T) - n^{\boldzero}_{i}(T) \right) \\
\labelrel{\geq}{i:7-c} \; &  \sum_{i=2}^{K} \Delta_i \times \left( \frac{\alpha \ln T}{\Delta_i^2} \left( \frac{2}{\Delta_i} \left[ \sqrt{\frac{\alpha \ln T}{n^{\boldzero}_1(T)}}  - \sqrt{\frac{\alpha \ln T}{s_1 + n^{\dd}_{1}(T)}} \right] - \frac{3}{\Delta_i^2} \frac{\alpha \ln T}{n^{\boldzero}_{1}(T)} \right) -  s_i \right) \\
= \; & \sum_{i=2}^{K} \frac{2 (\alpha \ln T)^{1.5}}{\Delta_i^2} \left[ \frac{1}{\sqrt{n^{\boldzero}_1(T)}}  - \frac{1}{\sqrt{s_1 + n^{\dd}_{1}(T)}} \right] - \Delta_i \; s_i - \frac{3 (\alpha \ln T)^2}{\Delta_i^3 \; n^{\boldzero}_{1}(T)}.
\end{align*}
Here, \eqref{i:7-c} uses \cref{clm:dd-subopt-lower,clm:zero-subopt-upper}.
\end{proof}

\section{One-shot comparison of $\ucb$ versus Thompson Sampling.}
\label{appn-sec:one-shot-comparison}
This section presents a minimal example illustrating a qualitative difference between $\ucb$ and Thompson Sampling in how they respond to side-information.
In this example, we show that providing a single extra sample of the optimal action decreases $\ucb$’s probability of selecting that optimal action in the final round, 
while the same side-information increases this probability for Thompson Sampling.
The example serves to contrast the distinct ways in which these algorithms incorporate side-information.

\UcbTsDifference*

The instance setup and the following \cref{lem:ucb-ts-diff-regs} complete the proof of the Theorem.

\paragraph{Instance setup.} Let time horizon $T=3$ and two actions $\actions = \{1,2\}$.
The rewards are Gaussian with unit variance and means
\[
\mu = \bigl(\tfrac{1}{2}+\Delta,\ \tfrac{1}{2}\bigr),
\]
for some $\Delta \in (0,\frac{1}{2}]$ so that action~$1$ is optimal.
We consider the side-information design $\dd$ defined by
\[
\dd_t = (0,0)\ \text{for } t=1,2,
\qquad
\dd_3 = (1,0),
\]
so that no side-information is provided initially, and a single additional sample of the optimal action is revealed at the beginning of the final time-step before the final action is taken.

Thompson Sampling (in short, $\ts$) is initialized with an improper flat prior belief $P_0(i) = \gaussian{0}{\infty}$ for $i \in \actions$.
Both algorithms pull each action once in the first two rounds in order to initialize their estimates and proper beliefs.
\\

\begin{lemma}
\label{lem:ucb-ts-diff-regs}
When the $\ucb$ and Thompson Sampling ($\ts$) algorithms run on with action set $\actions$ with and without side-information $\dd$ for horizon $T=3$, their expected regrets obey:
\begin{align*}
R(\ucb, (\actions, \boldzero), T) & \leq R(\ucb, (\actions, \dd), T) \quad \text{ and }\\
R(\ts, (\actions, \boldzero), T) & \geq R(\ts, (\actions, \dd), T).
\end{align*}
\end{lemma}

\begin{proof}
The Lemma is shown from the conclusions of 
\cref{appn-sec:one-shot-ucb,appn-sec:one-shot-ts} by explicitly working out the probabilities.
\end{proof}

\noindent We write $\probunder{\dd}{\cdot}$ for probabilities induced by running the algorithm under design $\dd$, and $\probunder{\boldzero}{\cdot}$ for the no-side-information design.
We write $\gaussian{\rho}{\sigma}$ to denote the guassian distribution with mean $\rho$ and variance $\sigma$, 
and write $\eta(\rho, \sigma)$ to denote a random variable that is distributed as $\gaussian{\rho}{\sigma}$.

After two rounds, both algorithms have observed exactly one sample from each action.
Thus,
\begin{align}
\muhat_2(1) \sim \gaussian{\mu(1)}{1},
\qquad
\muhat_2(2) \sim \gaussian{\mu(2)}{1}, \label{eqn:t2-estimates}  
\end{align}
and the joint distribution of observations up to $t=2$ is identical under $\dd$ and $\boldzero$. (The side-informations only differ at $t=3$.)

\subsection{$\ucb$ trajectory. }
\label{appn-sec:one-shot-ucb}
Denote $r(N,n)=\sqrt{\frac{2 \ln N}{n}}$ be the confidence radius of an action where where $n$ is the number of samples of an action and $N$ is the total number of samples observed so far over all actions.

\paragraph{No side-information.}
At the final step $t=3$, both actions have one sample.
The probability of playing the optimal action $1$ is 
\begin{align*}
\probunder{\boldzero}{x_3 = 1} &= \probunder{\boldzero}{\mathrm{UCB}_3(1) > \mathrm{UCB}_3(2)} = \probunder{\boldzero}{\muhat_2(1) + r(2,1) > \muhat_2(2) + r(2,1)} \\
& = \probunder{\boldzero}{\muhat_2(1) > \muhat_2(2)} = \prob{\eta(\mu(1),1) - \eta(\mu(2),1)} \labelrel{=}{i:9a} \prob{\eta(\Delta,2) > 0}  \\
& \labelrel{=}{i:9b} \Phi(\nicefrac{\Delta}{\sqrt{2}}) \geq \Phi(0.7 \times \Delta), \numberthis \label{eqn:ucb-prob-no-si}
\end{align*}
where \eqref{i:9a} uses the independence of action rewards, and thus the estimates,
and \eqref{i:9b} uses $\Phi$ as the cumulative density function of the standard normal distribution.

\paragraph{With side-information.}
Under design $\dd$, action~$1$ receives an additional sample before the final round. 
Thus, its empirical mean deviates from \cref{eqn:t2-estimates} to become 
$$\muhat_2(1) \sim \gaussian{\mu(2)}{\nicefrac{1}{2}}$$ having a tigher variance.
Then, at $t=3$, we have $n(1)=2$ and $n(2)=1$ yielding
\begin{align*}
\probunder{\dd}{x_3 = 1} =& \probunder{\dd}{\mathrm{UCB}_3(1) > \mathrm{UCB}_3(2)} = \probunder{\dd}{\muhat_2(1) + r(3,2) > \muhat_2(2) + r(3,1)} \\
=& \probunder{\dd}{\muhat_2(1) - \muhat_2(2) > r(3,1) - r(3,2)} = \prob{\eta(\mu(1),\nicefrac{1}{2}) - \eta(\mu(1),1)  > \sqrt{2 \ln 3} \left( 1-  \tfrac{1}{\sqrt{2}}  \right)} \\
\leq & \prob{\eta(\Delta, \nicefrac{3}{2}) > 0.44 }
\leq \Phi\left( (\Delta - 0.35)/ \sqrt{\nicefrac{3}{2}}  \right) \leq \Phi( 0.82 (\Delta - 0.35)). \numberthis \label{eqn:ucb-prob-w-si}
\end{align*}

\noindent Combining \cref{eqn:ucb-prob-no-si,eqn:ucb-prob-w-si}, we have that 
\begin{align}
\probunder{\dd}{x_3 = 1} \leq \Phi( 0.82 (\Delta - 0.35)) \labelrel{\leq}{i:9c} \Phi(0.7 \Delta) \leq \probunder{\boldzero}{x_3 = 1}, 
\end{align}
where \eqref{i:9c} holds for all $\Delta < 2.92$.
So, providing side-information decreases the probability that $\ucb$ selects the optimal action at the final round.

\subsection{Thompson Sampling trajectory}
\label{appn-sec:one-shot-ts}
Thompson Sampling maintains Gaussian posteriors
\[
P_t(i) = \gaussian{\muhat_t(i)}{\nicefrac{1}{n_t(i)}},
\]
where $n_t(i)$ counts all observations of action~$i$, including side-information.
At each round $t$, it independently samples $\tilde{\mu}_t(i) \sim P_{t-1}(i)$ for every $i$,
and plays the action with the largest sample, $\argmax_{i \in \actions} \tilde{\mu}_t(i)$.

After two rounds of action play, both actions have one observation, so
\begin{align*}
P_2(1) = \gaussian{\muhat_2(1)}{1} \qquad \text{ and } P_2(2) = \gaussian{\muhat_2(2)}{1}
\end{align*}
Marginally, using \cref{eqn:t2-estimates},
\begin{align}
P_2(1) \sim \gaussian{\mu(1)}{2} \qquad \text{ and } P_2(2) \sim \gaussian{\mu(2)}{2}.  \label{eqn:post-dist-no-side-info}
\end{align}

\paragraph{No side-information.}
The probability of playing the optimal action at the final step $t=3$ is
\begin{align*}
\probunder{\boldzero}{x_3 = 1} &= \probunder{\boldzero}{\tilde{\mu}_t(1) > \tilde{\mu}_t(2)} \labelrel{=}{i:9d}
\prob{\eta(\mu(1),2) > \eta(\mu(2),2) } \\
& \labelrel{=}{i:9e} \prob{\eta(\mu(1) - \mu(2), 4) > 0} 
= \prob{\eta(\Delta, 4) > 0} = \Phi(\nicefrac{\Delta}{2}). \numberthis \label{eqn:ts-prob-no-si}
\end{align*}
Here, \eqref{i:9d} uses \cref{eqn:post-dist-no-side-info}, and \eqref{i:9e} uses the independence of the posterior samples $\tilde{\mu}_{\cdot}(\cdot)$.
Here, $\Phi$ denotes the c.d.f. of the standard normal distribution.

\paragraph{With side-information.}
Under design $\dd$, action~$1$ has two observations while action~$2$ has one.
Thus, 
\begin{align*}
P_2(1) = \gaussian{\muhat_2(1)}{\nicefrac{1}{2}} \qquad \text{ and } P_2(2) = \gaussian{\muhat_2(2)}{1}.
\end{align*}
With the empirical means themselves distributed as 
$\muhat_2(1) = \gaussian{\mu(1)}{\nicefrac{1}{2}}$ (with one sample from side-information, and one from action play) and $\muhat_2(2) = \gaussian{\mu(2)}{1}$ (with one sample from action play), 
the distribution about the posterior belief distributions obey
\begin{align}
P_2(1) \sim \gaussian{\mu(1)}{1} \qquad \text{ and } P_2(2) \sim \gaussian{\mu(2)}{2}.  \label{eqn:post-dist-w-side-info}
\end{align}
Then, the probability of playing the optimal action at the final step $t=3$ is
\begin{align*}
\probunder{\dd}{x_3 = 1} & = \probunder{\dd}{\tilde{\mu}_t(1) > \tilde{\mu}_t(2)} \labelrel{=}{i:9f}
\prob{\eta(\mu(1),1) > \eta(\mu(2),2) } \\
& \labelrel{=}{i:9g} \prob{\eta(\mu(1) - \mu(2), 3) > 0} = \prob{\eta(\Delta, 3) > 0} = \Phi(\nicefrac{\Delta}{\sqrt{3}}). \numberthis \label{eqn:ts-prob-w-si}
\end{align*}
Here, \eqref{i:9f} uses \cref{eqn:post-dist-w-side-info}, and \eqref{i:9g} uses the independence of the posterior samples $\tilde{\mu}_{t}(\cdot)$. \\

\noindent Thus, from \cref{eqn:ts-prob-no-si,eqn:ts-prob-w-si}, we have that $\probunder{\dd}{x_3 = 1} > \probunder{\boldzero}{x_3 = 1}$.
we see that side-information increases the probability that Thompson Sampling selects the optimal action in the final round.

\section{Proof of \cref{thm:etc-monotone}}
\label{appn-sec:etc-monotone-k}
\paragraph{Algorithmic Description and Regret Decomposition.}
The Explore-Then-Commit ($\etc$) algorithm proceeds in two phases: an exploration phase of length $T' < T$ where all actions are uniformly played in a round-robin fashion.
Let $\muhat(i)$ be the empirical reward mean of action $i$ at the end of this exploration phase from both the action play outcomes and side-information samples.
Then, the algorithm chooses (or `commits' to) the empirical best action $\istar = \argmax_{i \in \actions} \muhat(i)$ and plays this $\istar$ for all rounds $T'+1, \dots, T$ thereafter.

The regret of $\etc$ naturally decomposes into that of these two phases: the regret during the exploration phase and the regret during commit phase after that.

During the exploration phase, $\etc$ deterministically plays all actions in a
round-robin fashion for the first $T' < T$ rounds. 
This phase is not adaptive, i.e., the choice of actions played does not depend on the action outcomes/rewards and side-information sample values received.
Hence the regrets incurred in the exploration phase under instances $(\actions,\dd)$ and $(\actions,\ee)$ are identical.

\subsection{Regret inequality in Commit Phase}

To show the Theorem, what remains to be shown is that the regrets in the commit phase $[T'+1, T]$ obey 
\begin{align}
R(\etc, (\actions, \ee), [T'+1,T]) \leq R(\etc, (\actions, \dd), [T'+1,T]),  \label{eqn:etc-monotone-commit-phase}
\end{align}
where side-information $\ee$ dominates $\dd$ (i.e., $\ee \succeq \dd$ as in \cref{dfn:side-info-dominance}).

We shall show this \cref{eqn:etc-monotone-commit-phase} in two steps: 
\begin{enumerate}
\item First, we show in \cref{clm:ee-less-var-than-dd} that the action reward mean estimates in instance $(\actions, \ee)$ with richer/dominating side-information `more accurate' than the estimate in instance $(\actions, \dd)$.
\item Second, with the estimates behaving as such, we show in \cref{lem:subopt-less-probable} that the probability of committing to a sub-optimal action in the commit phase is `lesser' when the learner has richer side-information.
\end{enumerate}

We setup some notations.
Write $\mu = (\mu(1), \mu(2), \dots, \mu(K)) \in \Real^K$ to be the list/vector of reward means of all actions in $\actions$,
and analogously define $\muhat = (\muhat(1), \dots, \muhat(K))$ to be the vector of empirical reward means.
We characterize the statistical properties of $\muhat$ under the two problem instances corresponding to the two side-information designs $\dd, \ee$.


We assume, as elsewhere in the paper, unit-variance Gaussian reward distributions, so that all reward and side-information samples of action $i$ are i.i.d.\ draws from $\gaussian{\mu(i)}{1}$.
Let $n(i)$ denote the number of times $\etc$ plays action $i$ during the (round-robin) exploration phase; since this phase is deterministic and unaffected by side information, $n(i)$ is the same under $\dd$ and $\ee$.
Let $s_\dd(i) := \sum_{u=1}^{T'} \dd_u(i)$ and $s_\ee(i) := \sum_{u=1}^{T'} \ee_u(i)$ denote the number of side-information samples of action $i$ revealed by the end of the exploration phase under designs $\dd$ and $\ee$ respectively; dominance $\ee \succeq \dd$ (\cref{eqn:side-info-dominance}) gives $s_\ee(i) \geq s_\dd(i)$ for every action $i$.

\begin{claim}[Dominant side-information $\Rightarrow$ Tighter estimate]
\label{clm:ee-less-var-than-dd}
The estimate $\muhat$ obeys the gaussian distribution with expectation and variance as follows:
\begin{align*}
    \muhat \sim_{\dd} \gaussian{\mu}{\Sigma_{\dd}}  \text{ and } \muhat \sim_{\ee} \gaussian{\mu}{\Sigma_{\ee}}, \text{ with } \Sigma_{\ee} \preceq \Sigma_{\dd},
\end{align*}
where $\Sigma_{\dd} = \mathrm{diag} \left( \frac{1}{n(1)+s_\dd(1)}, \dots, \frac{1}{n(K)+s_\dd(K)} \right)$, and similarly for $\Sigma_\ee$.
\end{claim}
\begin{proof}
Since the $n(i) + s_\dd(i)$ (resp. $n(i) + s_\ee(i)$) samples of action $i$ pooled by the commit time are i.i.d.\ draws from $\gaussian{\mu(i)}{1}$, and are independent across actions, the empirical mean vector $\muhat$ is gaussian with mean $\mu$ and the stated diagonal covariance under each design.
Since $s_\ee(i) \geq s_\dd(i)$ for every action $i$, we have $n(i) + s_\ee(i) \geq n(i) + s_\dd(i)$, so each diagonal entry of $\Sigma_\ee$ is at most the corresponding entry of $\Sigma_\dd$; as both covariances are diagonal, this gives $\Sigma_\ee \preceq \Sigma_\dd$.
\end{proof}

Without loss of generality, assume that actions in $\actions$ are numbered in decreasing optimality, i.e., $\mu(1) > \mu(2) > \cdots > \mu(K)$.

Let $\pd{\cdot}$ and $\pe{\cdot}$ be the probability measures of the algorithmic trajectory (or history) induced by running $\etc$ with on problem instances $(\actions, \dd)$ and $(\actions, \ee)$ respectively.

\begin{lemma}[Sub-optimal action play probabilities]
\label{lem:subopt-less-probable}
For any action $1 \leq i \leq K$, the probability of $\etc$ committing to action at least as sub-optimal as $i$ with richer side-information $\ee$ is at most the corresponding probability with side-information $\dd$. That is,
\begin{align*}
    \pe{\istar \in \{i, i+1, \dots, K\} } \leq \pd{\istar \in \{i, i+1, \dots, K \} }.
\end{align*}
\end{lemma}
\begin{proof}
Towards proving this, we develop some constructs using a generic side-information design $\ff$ and shall later instantiate them using appropriate side-information designs $\dd$ and $\ee$.

We introduce the following collection of sets $\{C_{A,i,t} \}_{i \in [K]}$, where we define
\begin{align}
C_{i} = \{\mu' \in \Real^K : \mu'(i) \geq \mu'(j) \text{ for all } j \neq i. \} \label{eqn:play-param-set}
\end{align}
to be the set/region of values that estimate vector $\muhat$ should belong to for action $i$ to be committed to by $\etc$.
Precisely, 
\begin{align}
    \muhat \in C_{i} \iff \istar = i, \label{eqn:action-prob-estimate-prob}
\end{align}
We shall study the probability that $\etc$ commits to a certain action $i$ by
studying the probability that the computed estimate vector $\muhat$ falls in $C_{i}$.

\begin{claim}[Convex cone]
\label{clm:convex-cone}
For any action $i \in [K]$, the set $C_{i}$ is a convex cone.
\end{claim}
\begin{proof}
The Claim follows since $C_i = \bigcap_{j \neq i} \{\mu' \in \Real^K : \mu'(i) - \mu'(j) \geq 0\}$ is an intersection of homogeneous half-spaces, each linear in $\mu'$ and hence itself a convex cone.

First, it is a cone since for all $\mu' \in C_{i}$, we have $c \mu' \in C_{i}$ for any positive constant $c>0$, as
\begin{align*}
\mu'(i) = \max_{j \in [K]} \mu'(j) \iff c \mu'(i) = \max_{j \in [K]} c \mu'(j).
\end{align*}

Second, it is convex since for any $\mu', \mu'' \in C_{i}$, we have that $c_1 \mu' + c_2 \mu'' \in C_{i}$ for positive constants $c_1, c_2 \geq 0$, as
\begin{align*}
(\mu'(i) = \max_{j} \mu'(j)) \land (\mu''(i) = \max_{j} \mu''(j)) & \implies (c_1. \mu'(i) + c_2 . \mu''(i) = c_1 . \max_{j} \mu'(j) + c_2 . \max_{j} \mu''(j)) \\
& \implies c_1. \mu'(i) + c_2 . \mu''(i) = \max_{j} c_1 .\mu'(j) + c_2 . \mu''(j)).
\end{align*}
\end{proof}
Geometrically, the collection $\{C_{i} \}_{i \in [K]}$ partitions $\Real^K$ into pie slices with apex/centre at origin.

We recall the following notion, relaxing convexity, along with a fact about Gaussian probabilities over such sets.

\begin{definition}[Star-shaped set]
\label{dfn:star-shaped-set}
Let $V$ be a Euclidean space. A set $S \subseteq V$ is said to be \emph{star-shaped} if there exists some point/element $x \in S$ such that for every $y \in S$,
the line segment $\overline{xy}$ lies within $S$, i.e.,
$\forall \lambda \in [0,1] : \lambda x + (1-\lambda) y \in S$.

The set of all such $x$ is called the \emph{kernel} of set $S$.
\end{definition}

\noindent A convex set is also star-shaped, and it can be seen that the entire convex set is its own kernel.

\noindent These star-shaped sets in the two-dimensional Euclidean plane are known as star-shaped polygons and appear to be well-studied in the computational geometry literature (see e.g., the 1985 textbook of \cite{preparata2012computational}).

\begin{claim}[Kernel-centred Gaussian distributions over star-shaped polytopes]
\label{clm:star-polytope-gaussian-probs}
Consider a star-shaped polytope $S$ with $\mu$ in its kernel. Let $A \sim \gaussian{\mu}{\Sigma_A}$ and $B \sim \gaussian{\mu}{\Sigma_B}$ be Gaussian random variables with $\Sigma_A \preceq \Sigma_B$, i.e., $A$ is more tightly centred around $\mu$ than $B$ is, in every direction.

Then, $\prob{A \in S} \geq \prob{B \in S}$.
\end{claim}
\begin{proof}
Write $f_A$ (and $f_B$) to be the p.d.f of $\gaussian{\mu}{\Sigma_A}$ (sim. $\gaussian{\mu}{\Sigma_B}$) in $d$ dimensions.
We show the Claim by comparing the corresponding probability integrals in the polar system.
We start with the L.H.S.
\begin{align*}
    \prob{A \in S} & = \int_{s \in S} f_A(s) .ds = \int_{s \in S}  (2 \pi)^{d/2} \det(\Sigma_A)^{-1/2} \exp{\frac{-1}{2} \norm{s - \mu}_{\Sigma_A^{-1}} } .ds \\
    & \labelrel{=}{i:10-a} \int_{\phi} \int_{r = 0}^{r_\phi} (2 \pi)^{d/2} \det(\Sigma_A)^{-1/2} \exp{\frac{-1}{2} \norm{r}_{\Sigma_A^{-1}} } . dr . d \phi \\
    & \labelrel{\geq}{i:10-b} \int_{\phi} \int_{r = 0}^{r_\phi} (2 \pi)^{d/2} \det(\Sigma_B)^{-1/2} \exp{\frac{-1}{2} \norm{r}_{\Sigma_B^{-1}} } . dr . d \phi \\
    & \labelrel{=}{i:10-c} \int_{s \in S} f_B(s) .ds = \prob{B \in S},
\end{align*}
where \eqref{i:10-b} is due to $\Sigma_A \preceq \Sigma_B$.
Further, crucially, the conversion to and from the polar systems in \eqref{i:10-a} and \eqref{i:10-c} over a continuous $[0,r_\phi]$ is possible because $S$ is star-shaped with $\mu$ in its kernel, i.e., in any direction $\phi$, a ray originating from $\mu$ exits the polytope $S$ at most once (and never enters again), and we call $r_\phi$ to be the distance of this exit point from $\mu$ (which is $\infty$ if it doesn't exit).
\end{proof}

To show the Lemma, as observed in \cref{eqn:action-prob-estimate-prob},
we shall show that, for all $1 \leq i \leq k$, $\muhat$ falls in the corresponding union of sets with a higher probability for richer side-information:
\begin{claim}
    $\pe{\muhat \in \bigcup_{j=1}^i C_{j}} \geq \pd{\muhat \in \bigcup_{j=1}^i C_{j}}$ for all $1 \leq i \leq K$.
\end{claim}
\begin{proof}
With the gaussian nature of distribution of $\muhat$ as in \cref{clm:ee-less-var-than-dd}, 
we are argue due to \cref{clm:star-polytope-gaussian-probs},
it is sufficient to show that the union of regions $\bigcup_{j=1}^i C_{i}$ is star-shaped polytope (\cref{dfn:star-shaped-set}) with its kernel containing the true $\mu$, the original reward means vector.

To start, for $i=1$, $C_{i}$ is a convex set (\cref{clm:convex-cone}) and is thus star-shaped, so the entire set is its own kernel.
Further $\mu \in C_1$ as if estimate $\muhat = \mu$ is accurate, then the algorithm commits to the optimal action $i=1$.
Thus, $\mu$ belongs to the kernel of $C_1$.

For $i \in [2,k]$, we show this using proof by contradiction.
Assume for some $i$ that $\bigcup_{j=1}^i C_{i}$ is not a star-shaped polytope w.r.t kernel point $\mu$.
Then, there exists some ray originating at $\mu$ that traverses $\mu^\ell$ and $\mu^g$ (in that order) such that 
$\mu^\ell \in C_{\ell}$ and $\mu^g \in C_{g}$ with $\ell > i \geq g$, i.e., action $g$ is more optimal than action $\ell$.
In other words, using observation \cref{eqn:action-prob-estimate-prob}, there is some direction in which as the estimate $\muhat$ moves away from the true $\mu$, the committing action of $\etc$ shall ceases to be the optimal action $1$ and changes to the sub-optimal action $ell$ as the estimate gets to $\mu^\ell$, and then changes again to a relatively optimal action $g$ as the estimate moves further away to $\mu^g$. 

As $\mu^\ell \in C_{\ell}$, by its definition \cref{eqn:play-param-set}, 
\begin{align*}
    & \mu^\ell(\ell) \geq \mu^\ell(g) \\
    \labelrel{\implies}{i:etc-ray-interpolation} & (c \mu(\ell) + (1-c) \mu^g(\ell)) \geq (c \mu(g) + (1-c) \mu^g(g)) \\
    \implies & c \mu(\ell) \geq c \mu(g) + (1-c) \left( \mu^g(g) - \mu^g(\ell) \right) \\
    \labelrel{\implies}{i:etc-ray-positive-gap} & c \mu(\ell) \geq c \mu(g) \numberthis \label{eqn:ray-contra}
\end{align*}
where \eqref{i:etc-ray-interpolation} is by linear interpolation for some $0 < c < 1$, \eqref{i:etc-ray-positive-gap} uses that $\mu^g(g) - \mu^g(\ell) > 0$.

Finally, \cref{eqn:ray-contra} implies that action $\ell$ is more optimal than action $g$, which is a contradiction.
This completes the proof of the Claim.
\end{proof}
The above Claim, in conjunction with \cref{eqn:action-prob-estimate-prob}, completes the proof of the Lemma.
\end{proof}

Finally, to complete the proof of the Theorem, we show in \cref{clm:insta-regret-inequality}, the instantaneous regret in the commit phase is lower with side-information $\ee$ than it is with $\dd$.

\begin{claim}[Instantaneous regret inequality]
\label{clm:insta-regret-inequality}
The per-time regret during the commit phase obeys:
\begin{align*}
    \sum_{j=2}^K \pe{\istar = j} \Delta(j) \leq \sum_{j=2}^K \pd{\istar = j} \Delta(j),     
\end{align*}
where $\Delta(j) := \mu(1) - \mu(j)$ is the sub-optimality `gap' of action $j \in [2,K]$.
\end{claim}
\begin{proof}

Write short-hands $p^{\ee}_i = \pe{\istar = i}$, $p^{\ee}_{i,K} = \pe{\istar \in \{i, i+1, \dots, K \}}$ and similarly define $p^{\dd}_i, p^{\dd}_{i,K}$.

We show the Claim by induction on action index $i$.  \paragraph{Hypothesis.} 
\begin{equation}
H(i): \sum_{j=i}^k \left( p^{\ee}_j - p^{\dd}_j \right) \Delta_j \leq \left( p^{\ee}_{i,K} - p^{\dd}_{i,K} \right) \Delta_i.
\end{equation}
\paragraph{Base Case.} Statement $H(k)$ holds as
\begin{align*}
    p^{\ee}_K \cdot \Delta_k - p^{\dd}_K \cdot \Delta_k = \left(p^{\dd}_{K,K} - p^{\ee}_{K,K} \right) \Delta_K.
\end{align*}

\paragraph{Induction Step.} Let $H(i+1)$ be true for some $i+1 \leq K$. We show $H(i)$ is true.
\begin{align*}
    \sum_{j=i}^{K} \left( p^{\ee}_j - p^{\dd}_j \right) \Delta_j = & \left( p^{\ee}_i - p^{\dd}_i \right) \Delta_i + \sum_{j=i+1}^{K} \left( p^{\ee}_j - p^{\dd}_j \right) \Delta_j \\
    \labelrel{\leq}{i:6-a} & \left( p^{\ee}_i - p^{\dd}_i \right) \Delta_i + \left( p^{\ee}_{i+1,K} - p^{\dd}_{i+1,K} \right) \Delta_{i+1} \\
    \labelrel{\leq}{i:6-b} & \left( p^{\ee}_i - p^{\dd}_i \right) \Delta_i + \left( p^{\ee}_{i+1,K} - p^S_{i+1,K} \right) \Delta_{i} = \left( p^{\ee}_{i,K} - p^{\dd}_{i,K} \right) \Delta_i. \numberthis \label{eqn:insta-reg-ineq}
\end{align*}
Here, \eqref{i:6-a} upper bounds the second term by using the induction assumption $H(i+1)$, 
then \eqref{i:6-b} is due to $p^{\ee}_{i+1,K} - p^{\dd}_{i+1,K} \leq 0$ by \cref{lem:subopt-less-probable} and $\Delta_i \leq \Delta_{i+1}$.
And \cref{eqn:insta-reg-ineq} shows $H(i)$. By mathematical induction, we have that $H(2)$ holds. \\

\noindent Along with the observation that $p^{\ee}_{i,K} - p^{\dd}_{i,K} \leq 0$ from \cref{lem:subopt-less-probable}, statement $H(2)$ implies the Claim.
\end{proof}
\section{Proof of \cref{thm:meta-algo-monotone}}
\label{appn-sec:meta-algo-proof}

We give a short Proof sketch, before restating the Theorem and giving the formal complete proof.
\paragraph{Proof sketch.}
Consider two side-information designs $\dd$ and $\ee$ such that $\ee \succeq \dd$.
The proof proceeds by coupling the executions of the meta-algorithm $\nondegalg$ under the two designs.

The key observation is that the internal trajectory of the black-box algorithm $\alg$ is identical in both runs.
Indeed, $\nondegalg$ only supplies side-information to $\alg$ at times when $\alg$ itself requests to play an action; from $\alg$’s perspective, receiving a reward sample from the environment or from the side-information buffer is indistinguishable.
Consequently, the sequence of actions recommended by $\alg$ is the same under $\dd$ and $\ee$.

Under this coupling, the only difference between the two runs is the availability and consumption of side-information.
By \cref{lem:e-runs-long-consumes-more-side-info}, dominance of $\ee$ over $\dd$ implies that, for every action $i$, the number of side-information samples consumed under $\ee$ is at least that under $\dd$, and hence the total number of consumed samples satisfies $\tau^\ee \ge \tau^\dd$.

The regret of $\nondegalg$ admits a decomposition into two parts: 
(i) the regret incurred by running the black-box algorithm $\alg$ for $T+\tau$ effective steps on the problem instance without side-information, and 
(ii) a correction term that subtracts the regret saved by replacing environment pulls with side-information samples. 
Comparing the two designs, the difference in regret reduces to the effect of extending the black-box run from $T+\tau^\dd$ to $T+\tau^\ee$ and the additional side-information consumed.

Finally, since $\alg$ satisfies \cref{assm:alg-plays-subopt-fewer}, the regret accumulated by $\alg$ over the additional window $[T+\tau^\dd,\, T+\tau^\ee]$ is at most proportional to the number of additional side-information samples consumed.
By the balanced consumption property of $\nondegalg$, this increase is exactly offset by the extra regret saved through those samples 
Therefore,
\[
R(\nondegalg,(\actions,\ee),T) \le R(\nondegalg,(\actions,\dd),T),
\]
establishing monotonicity of $\nondegalg$ with respect to dominating side-information.

We now restate the Theorem and then formally prove it.
\MetaAlgoMonotone*

\begin{proof}
Consider the action set $\actions$ and side-information designs $\dd$ and $\ee$ that obey $\ee \succeq \dd$ ($\ee$ dominates $\dd$).

We shall refer to the problem run $(\nondegalg, (\actions, \dd), T)$ (similarly $(\nondegalg, (\actions, \ee), T)$) loosely as run $\dd$ (similarly, run $\ee$).
Write $c^\dd(i)$ to be the $c(i)$ counter of the side-information samples of action $i$ consumed in problem run $\dd$.
At the end of time $T$, denote $\tau^\dd = \sum_{i} c^\dd(i)$ to be the total number of side-information samples consumed in scenario $\dd$.
The quantities are analogous defined for scenario $\ee$.
In general we shall use the superscripts $\dd,\ee$ to differentiate the algorithmic quantities w.r.t. side-information designs, 
and use subscripts (usually, $t,u,\rho$) to qualify the temporal information of the algorithmic quantities.

\begin{lemma}
\label{lem:e-runs-long-consumes-more-side-info}
The number of side-information samples consumed in scenario $\ee$ it at least as many as consumed in scenario $\dd$.
That is, $c^\ee(i) \geq c^\dd(i)$ throughout the time-horizon. 
Consequently, $\tau^\ee > \tau^\dd$ at the end of the time-horizon.
\end{lemma}
\begin{proof}
The black-box algorithm $\alg$'s trajectory is not affected by the side-information design.
This is because the meta-algorithm $\nondegalg$ delays the consumption of side-information such that it is fed to $\alg$ only when $\alg$ naturally requires to play that action.
From $\alg$'s point of view, it is identical to get the action outcome by playing the action in the environment and or to get it from the side-information buffer.

Let $\rho$ be used as a running variable to count the number of times the black-box $\alg$ has been queried.
Thus, when queried the $\rho$-th time, if the black-box $\alg$ recommends action $i$ under scenario $\dd$, then it shall also recommend action $i$ under scenario $\ee$.
Write $c^\ee_\rho(i), c^\dd_\rho(i)$ to be the side-information consumption counter values at the end of black-box algorithm's step $\rho$ in scenario $\ee$ and $\dd$ respectively. \\

We show the required result by induction over the black-box algorithm steps $\rho$ and the actions $i \in \actions$.

\noindent \textbf{Hypothesis. } For any step $\rho$, consider the hypothesis $H(\rho,i)$ : $c^\ee_{\rho}(i) \geq c^\dd_{\rho}(i)$.

\noindent \textbf{Base case. } Trivially, $c^\ee_0(i) = 0 \geq 0 = c^\dd_0(i)$ as no side-information is consumed before the meta-algorithm begins. Thus, $H(0,i)$ is true for all $i$.
\textbf{}{Induction step. }
Assume for some $\rho \in [T + \tau^\dd]$ that $H(\rho, i)$ is true for all $i$.

If $c^\dd_{\rho}(i) < c^\ee_{\rho}(i)$, then, as the couters are whole numbers, and as the counters can increase by at most one in a single step, it follows that $c^\dd_{\rho+1}(i) \leq c^\ee_{\rho+1}(i)$, thus $H(\rho, i)$ holds.

When $c^\dd_{\rho}(i) = c^\ee_{\rho}(i)$, we show $H(\rho+1, i)$ for each $i$ by arguing that 
when the black-box algorithm is queried for the $\rho+1$-th time,
if the meta-algorithm feeds it an outcome from the side-information buffer $S(i)$ in scenario $\dd$, then it does so in scenario $\ee$ as well, i.e., if $c^\dd(i)$ increases by one, then $C^\ee(i)$ shall also increase by one.

We show this by arguing about the two conditions in Line \ref{ln:condition-for-side-info} that if it is satisfied in scenario $\dd$, it is also satisfied in scenario $\ee$.
\begin{enumerate}
\item Let $t' \in [T]$ be the most recent environment time-step when the black-box algorithm is in internally in step $\rho+1$.
Always, all side information received by the meta-algorithm so far is either in the buffer or has been consumed.
Thus, 
writing $S^\ee_{\rho}(i) = \cardinality{S(i)}$ to be the cardinality of the buffer after step $\rho$, we have
$S^\ee_{\rho}(i) + c^\ee_{\rho}(i) = \sum_{u=1}^{t'} \ee_u(i)  \geq \sum_{u=1}^{t'} \ee_u(i)  = S^\dd_{\rho}(i) + c^\dd_{\rho}(i)$.
As $c^\ee_{\rho}(i) = c^\dd_{\rho}(i)$, we have that $S^\ee_{\rho}(i) \geq S^\dd_{\rho}(i)$, i.e., the buffer has in scenario $\ee$ at least as many samples as there is in scenario $\dd$.
Thus, if the buffer is note empty $S(i) \neq \emptyset$ in scenario $\dd$, then it is not empty in scenario $\ee$ also.
\item If $c^\dd_{\rho}(i) \leq \min_{j \in \actions} c^\dd_{\rho}(j)$, then,
\begin{align*}
    c^\ee_{\rho}(i) = c^\dd_{\rho}(i) \leq \min_{j \in \actions} c^\dd_{\rho}(j) \labelrel{\leq}{i:7d} \min_{j \in \actions} c^\ee_{\rho}(j),
\end{align*}
where \eqref{i:7d} uses the induction assumption $H(\rho,i)$ for all $i \in \actions$.
\end{enumerate}
The above two implications imply that if the meta-algorithm consumes an outcome from the side-information buffer $S(i)$ in scenario $\dd$, then it does so in scenario $\ee$ as well, i.e., if $c^\dd(i)$ increases by one, then $C^\ee(i)$ shall also increase by one.

By mathematical induction, we have that $c^\ee_{T+\tau^\dd}(i) \geq c^\dd_{T+\tau^\dd}(i)$ for all $i \in \actions$,
and thus $\tau^\ee \geq \tau^\dd$,
Further, as the counter value can only increase with the steps, $c^\ee_{T+\tau^\ee}(i) \geq c^\ee_{T+\tau^\dd}(i)$.
This completes the proof of the Lemma.
\end{proof}

Recall $\tau^\dd$ is the total number of side-information samples consumed.
Thus, the black box algorithm has been queried for a total of $T+\tau^\dd$ times.
Owing to the identically distributed nature of the rewards and side-information samples of an action, 
in effect,
the black box algorithm has been run on problem instance without side-informaton, 
i.e., $(\actions, 0)$ for a period of $T + \tau^\dd$ time-steps.
Write the regret of this black-box algorithm here to be $R(\alg, (\actions, 0), T+ \tau^\dd)$.

The meta-algorithm has incurred the regret of the black-box algorithm minus whatever samples it used from the side-information buffers to feed the black-box algorithm. 
Note that whenever it couldn't use samples from the side-information buffers, it played the action in the environment and incurred regret.
Thus, we have 
\begin{align}
R(\nondegalg, (\actions, \dd), T) = R(\alg, (\actions, 0), T + \tau^\dd) - \sum_{i \in \actions} c^\dd(i) \Delta_i. \label{eqn:reg-decompose-D} \\
R(\nondegalg, (\actions, \ee), T) = R(\alg, (\actions, 0), T + \tau^\ee) - \sum_{i \in \actions} c^\ee(i) \Delta_i.  \label{eqn:reg-decompose-E}
\end{align}
Let $T > T'$ be suitably large where the consumptions counts $c^\dd(i)$s are equal as per the balanced consumption of side-information.

We have
\begin{align*}
& R(\alg, (\actions, 0), T + \tau^\ee) - R(\alg, (\actions, 0), T + \tau^\dd) \\
\labelrel{=}{i:7aa} & R(\alg, (\actions, 0), [T + \tau^\dd, T + \tau^\ee]) \\
\labelrel{\leq}{i:7a} & \frac{\tau^\ee-\tau^\dd}{K} \sum_{i \in \actions} \Delta_i \\
\labelrel{=}{i:7b} & \sum_{i \in \actions} c^\ee(i) - c^\dd(i) \numberthis \label{eqn:black-box-window-regret} \\
\end{align*}
Here, \eqref{i:7aa} uses \cref{lem:e-runs-long-consumes-more-side-info} to have $[T+\tau^\dd, T+\tau^\ee]$ is a valid positive-sized time interval.
Then, \eqref{i:7a} uses that $\alg$ obeys \cref{assm:alg-plays-subopt-fewer}, and \eqref{i:7b} follows from the meta algorithm's balanced-consumption of side-information.
Now, substituting \cref{eqn:black-box-window-regret} into the equation obtained by subtracting \cref{eqn:reg-decompose-D} from \cref{eqn:reg-decompose-E}, we have
\begin{align*}
& R(\nondegalg, (\actions, \ee), T) - R(\nondegalg, (\actions, \dd), T) \\
= & R(\alg, (\actions, 0), T + \tau^\ee) - R(\alg, (\actions, 0), T + \tau^\dd) - \left( \sum_{i \in \actions} c^\ee(i) \Delta_i - \sum_{i \in \actions} c^\dd(i) \Delta_i \right) \\
\labelrel{\leq}{i:7c} &  0, \numberthis \label{eqn:algo-reg-ineq-dom-si}
\end{align*}
where \eqref{i:7c} is due to \cref{eqn:black-box-window-regret}.
This \cref{eqn:algo-reg-ineq-dom-si} shows the Theorem.
\end{proof}

\section{Additional Simulation Experiments}

\subsection{The argument about front-loading optimal action samples.}
\begin{figure}
\centering
\begin{subfigure}{.48\linewidth}
    \centering
    \includegraphics[width=\linewidth]{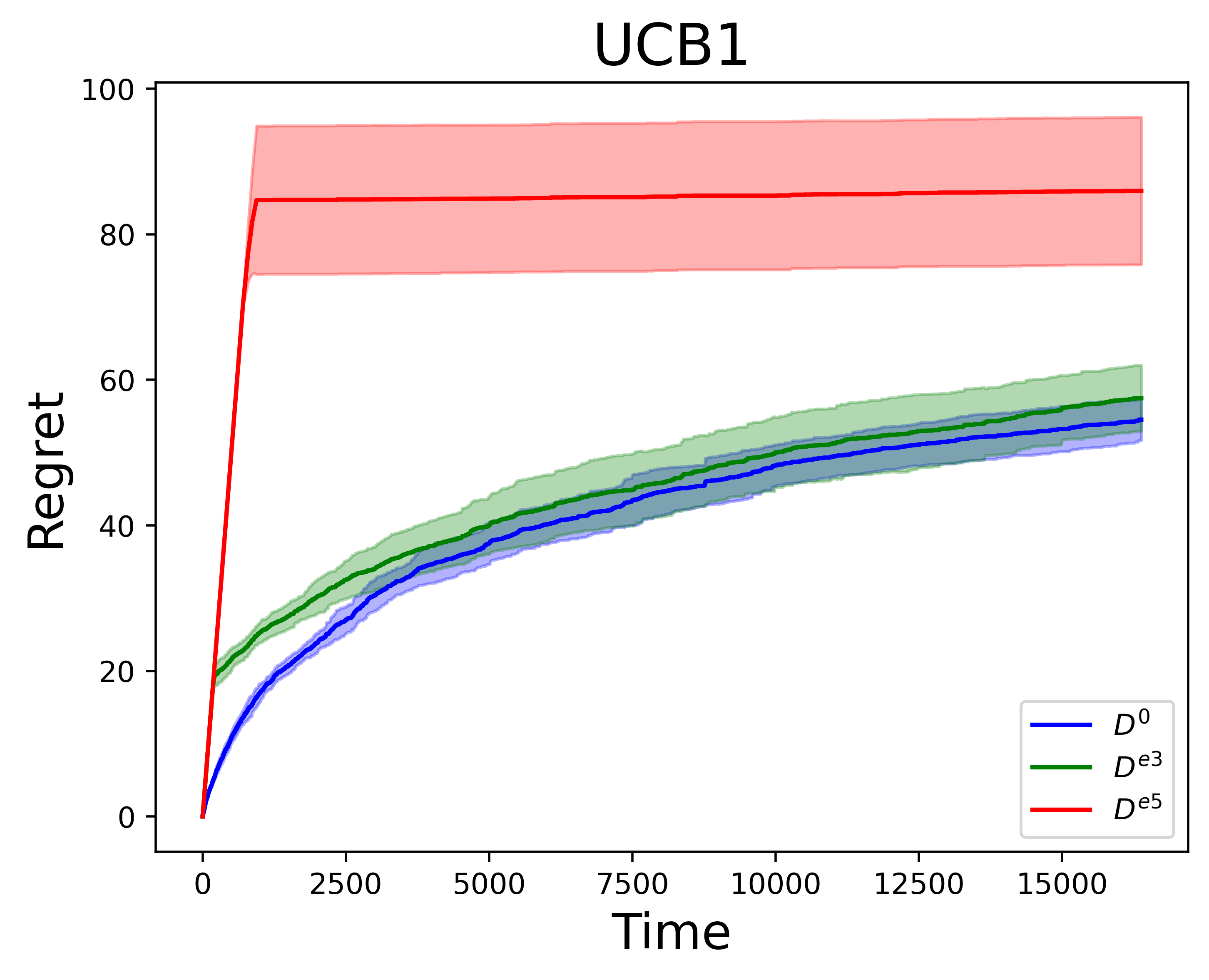}
    \caption{Impact of front-loading side-information on $\ucb$}\label{fig:overload-ucb}
\end{subfigure}%
\begin{subfigure}{.47\linewidth}
    \centering
    \includegraphics[width=\linewidth]{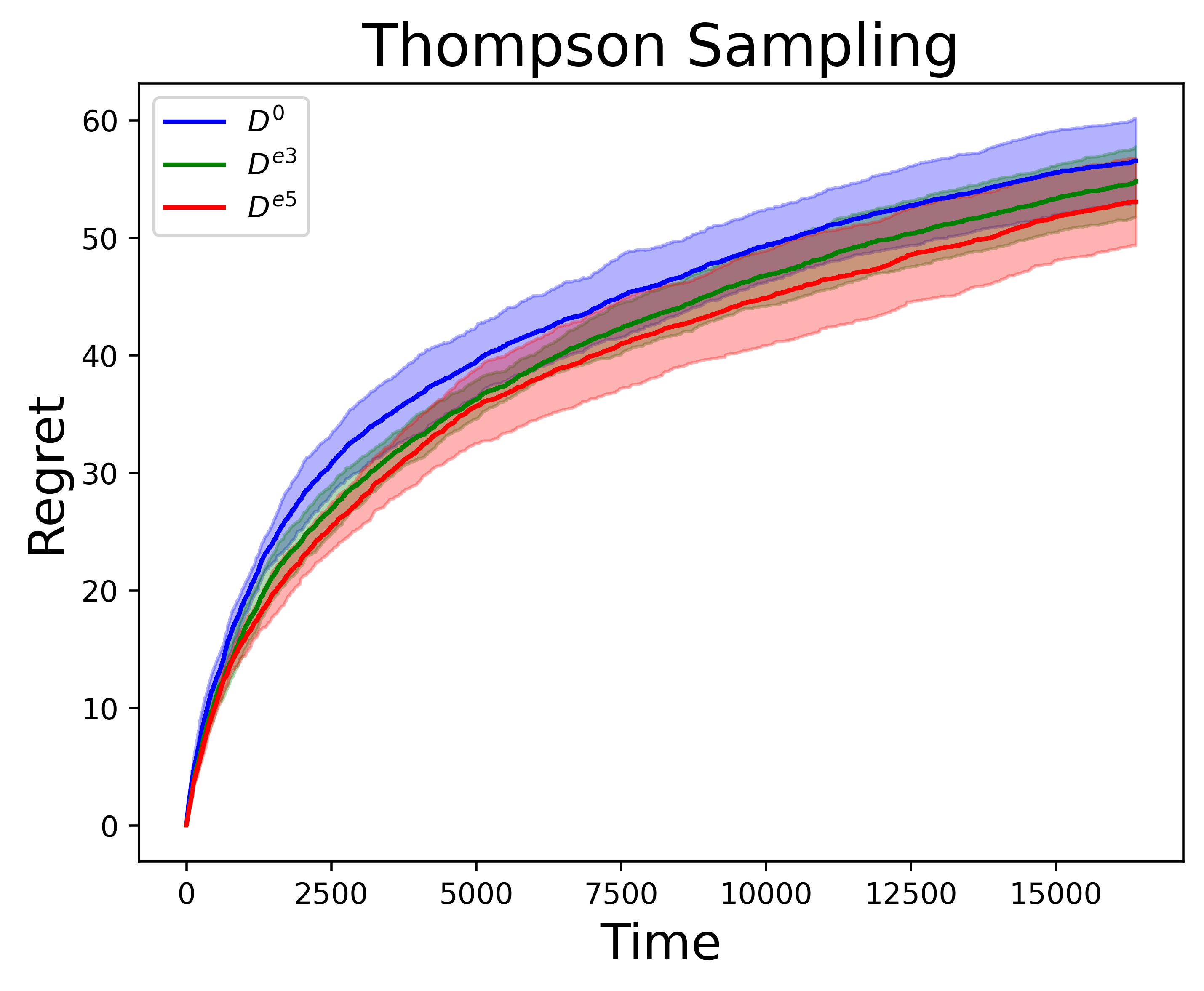}
    \caption{Impact of different agents on agent $g=6$.} \label{fig:overload-ts}
\end{subfigure}
\caption{For an instance of $\mu(\frac{1}{2}, \frac{1}{2} + \Delta)$, we show the regrets over three different side-information designs. $\dd^0$ contains no side-information. 
$\dd^{e3}$ and $\dd^{e5}$ provide 1,000 and 100,000 samples of the optimal action at time $t=0$.}
\label{fig:front-load}
\end{figure}

In our theoretical results (especially \cref{thm:ucb-degrades}), we employed instances where the optimal action's outcomes are provided at the begining as extra information. 
We argued that this leads to $\ucb$ incurs more regret, whereas Thompson Sampling doesn't.
We provide experimental backing to that in \cref{fig:front-load}.
The `catch-up' effect described in the main paper is evident.
Upon receiving huge amounts of optimal samples, $\ucb$ ends up exploring the sub-optimal action to a commensurate level by continuously playing the sub-optimal action at the beginning.
Whereas, Thompson Sampling doesn't suffer that. It starts playing the sub-optimal action less frequently right from the outset given the better estimate of the optimal action with more side-information.

\newpage
\section*{NeurIPS Paper Checklist}

The checklist is designed to encourage best practices for responsible machine learning research, addressing issues of reproducibility, transparency, research ethics, and societal impact. Do not remove the checklist: {\bf The papers not including the checklist will be desk rejected.} The checklist should follow the references and follow the (optional) supplemental material.  The checklist does NOT count towards the page
limit. 

Please read the checklist guidelines carefully for information on how to answer these questions. For each question in the checklist:
\begin{itemize}
    \item You should answer \answerYes{}, \answerNo{}, or \answerNA{}.
    \item \answerNA{} means either that the question is Not Applicable for that particular paper or the relevant information is Not Available.
    \item Please provide a short (1--2 sentence) justification right after your answer (even for \answerNA). 
\end{itemize}

{\bf The checklist answers are an integral part of your paper submission.} They are visible to the reviewers, area chairs, senior area chairs, and ethics reviewers. You will also be asked to include it (after eventual revisions) with the final version of your paper, and its final version will be published with the paper.

The reviewers of your paper will be asked to use the checklist as one of the factors in their evaluation. While \answerYes{} is generally preferable to \answerNo{}, it is perfectly acceptable to answer \answerNo{} provided a proper justification is given (e.g., error bars are not reported because it would be too computationally expensive'' or ``we were unable to find the license for the dataset we used''). In general, answering \answerNo{} or \answerNA{} is not grounds for rejection. While the questions are phrased in a binary way, we acknowledge that the true answer is often more nuanced, so please just use your best judgment and write a justification to elaborate. All supporting evidence can appear either in the main paper or the supplemental material, provided in appendix. If you answer \answerYes{} to a question, in the justification please point to the section(s) where related material for the question can be found.

IMPORTANT, please:
\begin{itemize}
    \item {\bf Delete this instruction block, but keep the section heading ``NeurIPS Paper Checklist"},
    \item  {\bf Keep the checklist subsection headings, questions/answers and guidelines below.}
    \item {\bf Do not modify the questions and only use the provided macros for your answers}.
\end{itemize}


\begin{enumerate}

\item {\bf Claims}
    \item[] Question: Do the main claims made in the abstract and introduction accurately reflect the paper's contributions and scope?
    \item[] Answer: \answerYes{} 
    \item[] Justification: \answerNA{}

\item {\bf Limitations}
    \item[] Question: Does the paper discuss the limitations of the work performed by the authors?
    \item[] Answer: \answerYes{} 
    \item[] Justification: \answerNA{}

\item {\bf Theory assumptions and proofs}
    \item[] Question: For each theoretical result, does the paper provide the full set of assumptions and a complete (and correct) proof?
    \item[] Answer: \answerYes{} 
    \item[] Justification: \answerNA{}

    \item {\bf Experimental result reproducibility}
    \item[] Question: Does the paper fully disclose all the information needed to reproduce the main experimental results of the paper to the extent that it affects the main claims and/or conclusions of the paper (regardless of whether the code and data are provided or not)?
    \item[] Answer: \answerYes{} 
    \item[] Justification: \answerYes{}

\item {\bf Open access to data and code}
    \item[] Question: Does the paper provide open access to the data and code, with sufficient instructions to faithfully reproduce the main experimental results, as described in supplemental material?
    \item[] Answer: \answerNo{} 
    \item[] Justification: \answerNA{}

\item {\bf Experimental setting/details}
    \item[] Question: Does the paper specify all the training and test details (e.g., data splits, hyperparameters, how they were chosen, type of optimizer) necessary to understand the results?
    \item[] Answer: \answerYes{} 
    \item[] Justification: \answerNA{}

\item {\bf Experiment statistical significance}
    \item[] Question: Does the paper report error bars suitably and correctly defined or other appropriate information about the statistical significance of the experiments?
    \item[] Answer: \answerYes{} 
    \item[] Justification: \answerNA{}

\item {\bf Experiments compute resources}
    \item[] Question: For each experiment, does the paper provide sufficient information on the computer resources (type of compute workers, memory, time of execution) needed to reproduce the experiments?
    \item[] Answer: \answerYes{} 
    \item[] Justification: \answerNA{}
    
\item {\bf Code of ethics}
    \item[] Question: Does the research conducted in the paper conform, in every respect, with the NeurIPS Code of Ethics \url{https://neurips.cc/public/EthicsGuidelines}?
    \item[] Answer: \answerYes{} 
    \item[] Justification: \answerNA{}

\item {\bf Broader impacts}
    \item[] Question: Does the paper discuss both potential positive societal impacts and negative societal impacts of the work performed?
    \item[] Answer: \answerNA{} 
    \item[] Justification: \answerNA{}
    
\item {\bf Safeguards}
    \item[] Question: Does the paper describe safeguards that have been put in place for responsible release of data or models that have a high risk for misuse (e.g., pre-trained language models, image generators, or scraped datasets)?
    \item[] Answer: \answerNA{} 
    \item[] Justification: \answerNA{}

\item {\bf Licenses for existing assets}
    \item[] Question: Are the creators or original owners of assets (e.g., code, data, models), used in the paper, properly credited and are the license and terms of use explicitly mentioned and properly respected?
    \item[] Answer: \answerNA{} 
    \item[] Justification: \answerNA{}

\item {\bf New assets}
    \item[] Question: Are new assets introduced in the paper well documented and is the documentation provided alongside the assets?
    \item[] Answer: \answerNA{} 
    \item[] Justification: \answerNA{}

\item {\bf Crowdsourcing and research with human subjects}
    \item[] Question: For crowdsourcing experiments and research with human subjects, does the paper include the full text of instructions given to participants and screenshots, if applicable, as well as details about compensation (if any)? 
    \item[] Answer: \answerNA{} 
    \item[] Justification: \answerNA{}

\item {\bf Institutional review board (IRB) approvals or equivalent for research with human subjects}
    \item[] Question: Does the paper describe potential risks incurred by study participants, whether such risks were disclosed to the subjects, and whether Institutional Review Board (IRB) approvals (or an equivalent approval/review based on the requirements of your country or institution) were obtained?
    \item[] Answer: \answerNA{} 
    \item[] Justification: \answerNA{}

\item {\bf Declaration of LLM usage}
    \item[] Question: Does the paper describe the usage of LLMs if it is an important, original, or non-standard component of the core methods in this research? Note that if the LLM is used only for writing, editing, or formatting purposes and does \emph{not} impact the core methodology, scientific rigor, or originality of the research, declaration is not required.
    \item[] Answer: \answerNA{} 
    \item[] Justification: \answerNA{}

\end{enumerate}

\end{document}